%% file: arXiv.tex
\documentclass{article}
\input{style}

\usepackage[eandd, final]{neurips_2026}

\usepackage[utf8]{inputenc} 
\usepackage[T1]{fontenc}    
\usepackage{hyperref}       
\usepackage{url}            
\usepackage{booktabs}       
\usepackage{multirow}       
\usepackage{amsfonts}       
\usepackage{nicefrac}       
\usepackage{microtype}      
\usepackage{xcolor}         

\title{How Private is Private? \\A Comparative Study for Face De-Identification}

\author{
  Hui Wei$^{1,2}$ \quad\quad Hao Yu$^{1,2}$ \quad\quad Hui Kuurila-Zhang$^{2}$ \quad\quad Guoying Zhao$^{1,2}$\thanks{Corresponding author}\\ 
  $^{1}$ELLIS Institute Finland, Finland \\
  $^{2}$Center for Machine Vision and Signal Analysis (CMVS), University of Oulu, Finland \\
  \texttt{\{hui.wei, hao.2.yu, hui.zhang, guoying.zhao\}@oulu.fi} \\
  \textcolor{magenta}{\href{https://cv-ac.github.io/hifd/}{https://cv-ac.github.io/hifd/}}
}

\begin{document}

\maketitle

\begin{abstract}
Face de-identification (FDeID) has emerged as a critical privacy-preserving technology, yet its evaluation remains fundamentally fragmented.
Existing protocols rely on inconsistent metrics, heterogeneous datasets, and partial annotation coverage, so methods targeting different utility dimensions, such as landmark versus expression preservation, are reported on different benchmarks under different metrics, rendering cross-method comparison infeasible.
We revisit FDeID evaluation from both the data and metric perspectives. On the data side, we introduce \uface, a curated, demographically balanced benchmark with high identity diversity, assembled from four large-scale face datasets through identity-aware cleaning, resolution enhancement, and stratified filtering. 
On the metric side, we propose \metric, a \textbf{Hi}erarchical \textbf{F}ace \textbf{D}e-identification metric that unifies identity suppression, multi-level utility preservation, and image quality under a single \emph{consistency-based} paradigm: every component is computed from pretrained estimators' outputs on the original face and its de-identified counterpart, directly quantifying how much identity is suppressed and how much downstream-perceivable utility survives.
\metric\ organizes facial signals into a three-level utility hierarchy spanning macro cues (L1), micro cues (L2), and imperceptible cues (L3), and aggregates the five resulting components into a single interpretable score via weighted harmonic mean, with configurable application-specific profiles.
Using this unified protocol, we conduct a comprehensive comparative study spanning adversarial, GAN-based, and diffusion-based methods, surfacing trade-offs and failure modes that remain invisible under existing protocols.
We release the benchmark and evaluation toolkit to foster systematic and reproducible research in privacy-preserving human face analysis.

\end{abstract}

\section{Introduction}
The ubiquity of visual sensing in modern society has intensified the tension between advancing computer vision capabilities and privacy protection~\cite{hukkelaas2019deepprivacy,hasan2023presentation,gu2024federated,jia2025balancing,laishram2025toward}.
Among all visual cues, human faces constitute uniquely identifying biometric signals and are captured pervasively by surveillance cameras, autonomous vehicles, and smart devices, often without explicit user consent~\cite{schroff2015facenet,deng2020retinaface}.
This reality has motivated extensive research on face de-identification (FDeID), which aims to modify facial imagery such that identity recognition is prevented while other task-relevant visual information is preserved for downstream applications~\cite{maximov2020ciagan,lopez2024privacy,hinojosa2022privhar,zhou2025improving}.
Despite over two decades of active research, a fundamental question remains insufficiently answered: \textbf{how private is private?}

\begin{figure}[!t]
  \centering
  \includegraphics[width=\linewidth]{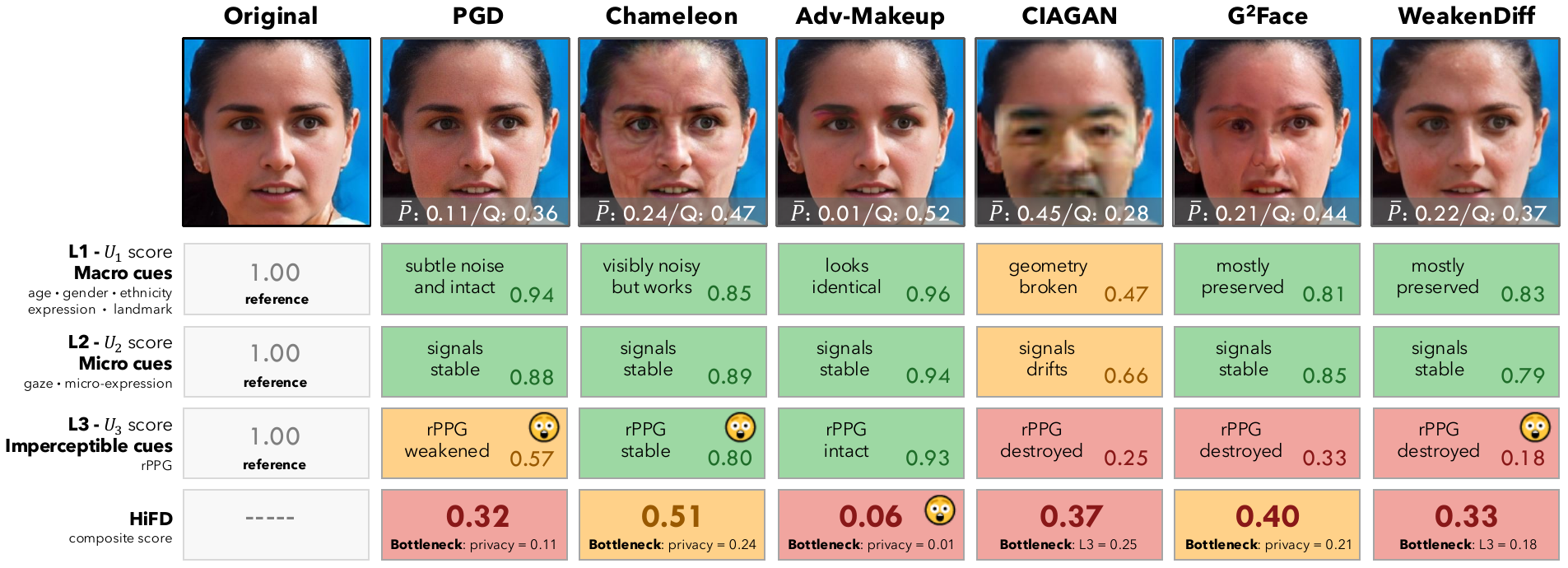}
  \vspace{-5mm}
  \caption{\textbf{\metric\ reveals what the eye and single-axis metrics cannot.} One face processed by six FDeID methods. Privacy score ($\bar{P}$) and quality score ($Q$) overlay each de-identified image. Rows report utility scores at each hierarchy level (L1~macro: $U_1$, L2~micro: $U_2$, L3~imperceptible: $U_3$) and the composite \metric. The \emph{bottleneck} tag names the weakest of $\{\bar{P}, Q, U_1, U_2, U_3\}$. Cells marked \raisebox{-0.2ex}{\includegraphics[height=1.8ex]{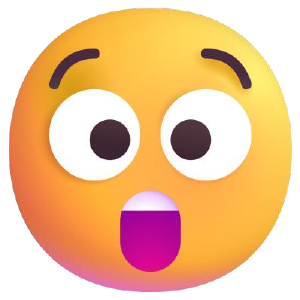}} flag scores that contradict naive intuition: (i)~Chameleon~\cite{chow2024personalized} looks visibly noisy yet $U_3=0.80$ because the rPPG signal remains stable, whereas PGD, whose perturbation is invisible, drops to $0.57$. (ii)~Adv-Makeup~\cite{yin2021adv} looks identical to the original yet collapses to $\metric=0.06$. (iii)~WeakenDiff~\cite{salar2025enhancing} passes L1 and L2 yet destroys the rPPG signal at L3.}
  \vspace{-7mm}
\label{fig:teaser}
\end{figure}

Answering this question is hard because FDeID is inherently multi-objective. A successful method must not only suppress identity but also preserve various categories of facial utility, \eg, age~\cite{geng2007automatic}, gender~\cite{meng2014face}, expression~\cite{pan2024opticaldr}, and landmarks~\cite{yang2024g}, with the relevant subset depending on the downstream application. The community has responded by developing methods that each target a different subset of these utilities, and by evaluating each method on whichever dataset happens to carry the corresponding annotations. The result is a fragmented evaluation landscape in which a method optimized for expression preservation~\cite{yang2024g} cannot be directly compared with one designed for detection preservation~\cite{jiang2023dartblur}, since they target fundamentally different utility dimensions on entirely different benchmarks~\cite{liu20253d,wen2023divide}. 
This fragmentation is reinforced by the absence of a unified benchmark: no existing dataset jointly provides the annotation coverage, identity diversity, and image quality required for reliable multi-dimensional evaluation~\cite{huang2008labeled,liu2015deep,moschoglou2017agedb,karras2019style}, so prior works have relied on heterogeneous task-specific datasets that produce mutually incompatible results. 
For instance, A$^3$GAN~\cite{zhai2022a3gan} evaluates expression preservation on ExpW~\cite{zhang2018facial}, whereas SASMask~\cite{gong2023stealthy} assesses age preservation on AgeDB~\cite{moschoglou2017agedb}, and the two simply cannot be meaningfully compared. We reviewed over 100 FDeID-related papers and quantified this fragmentation: more than \textit{60 datasets} and \textit{20 evaluation metrics} are used across prior protocols.

\begin{wrapfigure}{r}{0.55\linewidth}
  \centering
  \vspace{-5mm}
  \includegraphics[width=1.0\linewidth]{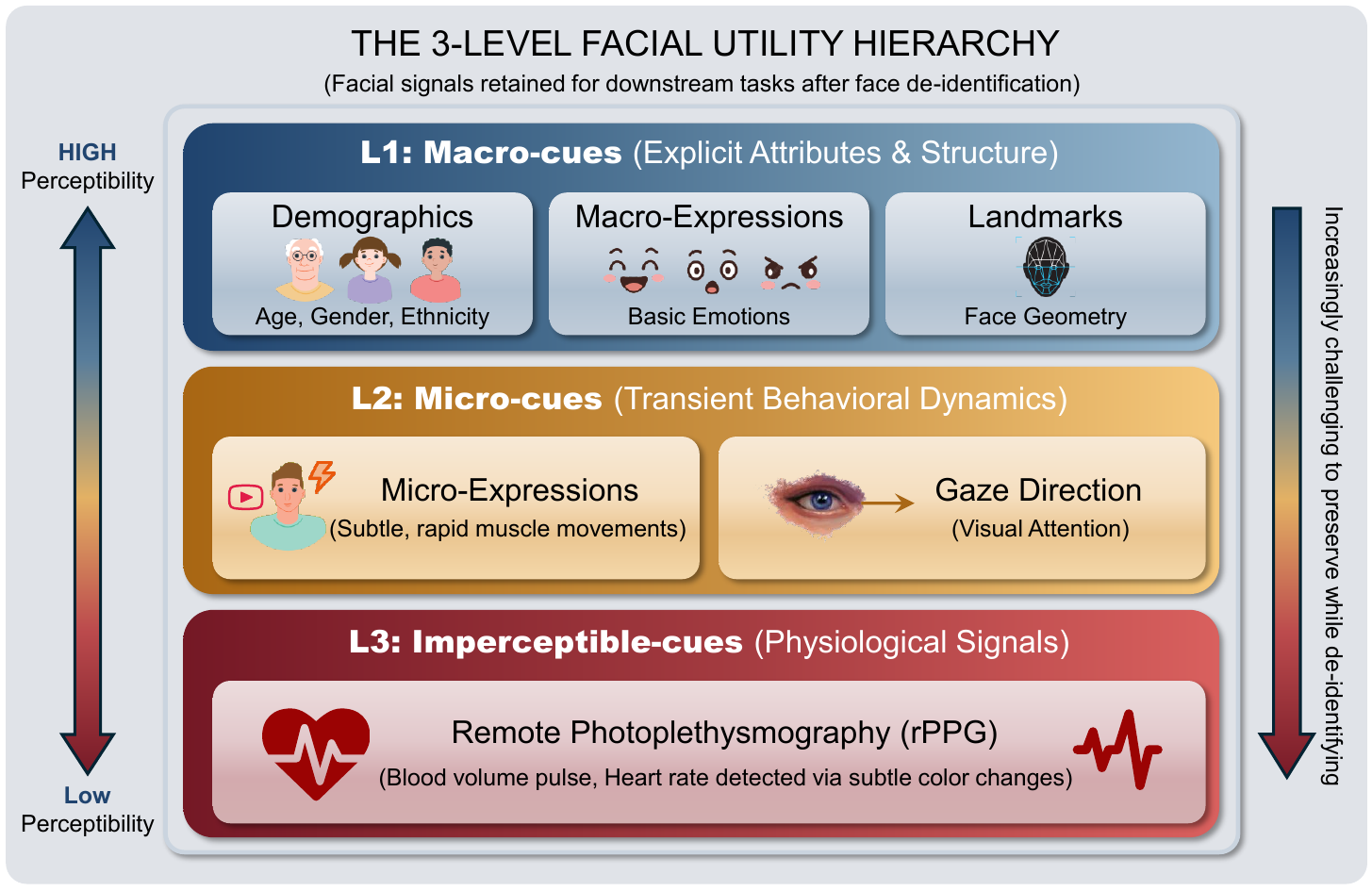}
  \vspace{-5mm}
  \caption{\textbf{The proposed 3-level facial utility hierarchy.} Facial utilities are categorized by perceptual accessibility into macro-cues (L1), micro-cues (L2), and imperceptible-cues (L3), motivating a hierarchical metric for FDeID evaluation.}
  \label{fig:hierarchy}
  \vspace{-3mm}
\end{wrapfigure}

Beyond the fragmentation of \emph{what} is measured, we identify a deeper concern in \emph{how} utility is measured. Existing protocols cast utility preservation as the classification or regression accuracy of attribute predictions against externally annotated labels: a method is said to preserve expression if a classifier still predicts the annotated expression label on the de-identified image~\cite{yang2021towards,salar2025enhancing}. This formulation is problematic on two counts. \emph{Operationally}, it requires per-attribute ground-truth annotations on the evaluation dataset, which is precisely why no single benchmark suffices for multi-attribute evaluation (Tab.~\ref{tab:comparison}) and why the field has fragmented across task-specific datasets in the first place. \emph{Conceptually}, it misaligns evaluation with deployment: what matters in practice is whether downstream models can still extract task-relevant information from the de-identified face, not whether their predictions match a static set of human annotations. We argue instead that utility should be measured by the consistency $s(f(\mathbf{x}),\, f(\hat{\mathbf{x}}))$ between an attribute estimator's predictions on the original face $\mathbf{x}$ and its de-identified counterpart $\hat{\mathbf{x}}$. This \textit{consistency-based} formulation directly quantifies whether the information a model extracts from a face survives de-identification, eliminates the per-image annotation requirement, and unifies evaluation across attributes and datasets.

Orthogonal to how utility is measured, we further observe that facial signals exhibit a natural hierarchy of perceptual granularity that existing protocols overlook (Fig.~\ref{fig:hierarchy}). Macro-level cues such as demographics and facial geometry are spatially coarse, semantically salient, and readily perceivable by human observers~\cite{moschoglou2017agedb,mollahosseini2017affectnet}. Micro-level cues such as gaze direction and micro-expressions depend on precise localized geometry and temporal dynamics that are far more fragile under facial manipulation~\cite{krafka2016eye,zhao2023dfme}. Imperceptible cues such as rPPG-based physiological signals, encoded in sub-pixel skin color variations, are the most susceptible to any image modification~\cite{stricker2014non,joshi2024factorizephys}. This stratification suggests that FDeID methods affect each level differently, which a single scalar utility score cannot expose, motivating a hierarchical metric that evaluates preservation at all three granularities and surfaces the trade-offs between them.

Based on the above observations, we make the following contributions:
\begin{itemize}[nosep,leftmargin=*]
    \item \textbf{\uface: a curated, balanced FDeID benchmark.} We construct a face benchmark assembled from multiple large-scale datasets through identity-aware cleaning, resolution enhancement, and stratified filtering, providing a demographically balanced evaluation substrate with high identity diversity and image quality. We further integrate established video benchmarks (DFME~\cite{zhao2023dfme}, PURE~\cite{stricker2014non}) to enable standardized evaluation across all utility levels.

    \item \textbf{A three-level facial utility hierarchy.} We propose the first structured formalization of facial utility for FDeID, organizing signals into macro-cues (L1), micro-cues (L2), and imperceptible-cues (L3), and providing a principled lens for understanding which downstream tasks remain viable after de-identification (see Fig.~\ref{fig:hierarchy}).

    \item \textbf{\metric: a unified, consistency-based metric.} We introduce \metric\ (\textbf{Hi}erarchical \textbf{F}ace \textbf{D}e-identification metric), which integrates identity suppression, multi-level utility preservation, and image quality into an interpretable score via weighted harmonic mean aggregation, with application-specific weighting profiles and graceful adaptation across image and video modalities (see Fig.~\ref{fig:teaser}).

    \item \textbf{A comprehensive comparative study of FDeID methods.} We evaluate representative adversarial, GAN-based, and diffusion-based methods under \uface\ and \metric, surfacing trade-offs and failure modes that remain invisible under fragmented protocols.
\end{itemize}

\setlength{\tabcolsep}{0.5em}
\begin{table*}[t]
\centering
\caption{\textbf{Annotation coverage of face datasets previously used for FDeID evaluation.} No prior dataset provides the joint L1+L2+L3 annotation coverage that fragmented FDeID protocols have collectively required, forcing each method to be evaluated on a different subset.} 
\vspace{-2mm}
\label{tab:comparison}
\resizebox{\textwidth}{!}{%
\begin{tabular}{@{}lcccccccccccc@{}}
\toprule
\multirow{2}{*}{\textbf{Dataset}} & \multirow{2}{*}{\textbf{Modality}} & \multirow{2}{*}{\textbf{\#Images/Videos}} & \multirow{2}{*}{\textbf{\#IDs}} & \multicolumn{5}{c}{\textbf{L1: Macro-Cues}} & \multicolumn{2}{c}{\textbf{L2: Micro-Cues}} & \textbf{L3} \\
\cmidrule(lr){5-9} \cmidrule(lr){10-11} \cmidrule(lr){12-12}
 & & & & \textbf{Age} & \textbf{Gender} & \textbf{Ethnicity} & \textbf{Macro-Exp.} & \textbf{Landmark} & \textbf{Micro-Exp.} & \textbf{Gaze} & \textbf{rPPG}  \\
\midrule
LFW~\cite{huang2008labeled}           & Image & 13,233   & 5,749   & \textcolor{red}{\xmark} & \textcolor{red}{\xmark} & \textcolor{red}{\xmark} & \textcolor{red}{\xmark} & \textcolor{red}{\xmark} & \textcolor{red}{\xmark} & \textcolor{red}{\xmark} & \textcolor{red}{\xmark} \\
CelebA~\cite{liu2015deep}     & Image & 202,599  & 10,177  & \textcolor{red}{\xmark} & \textcolor{green!60!black}{\cmark} & \textcolor{red}{\xmark} & \textcolor{green!60!black}{\cmark} & \textcolor{green!60!black}{\cmark} & \textcolor{red}{\xmark} & \textcolor{red}{\xmark} & \textcolor{red}{\xmark}  \\
FFHQ~\cite{karras2019style}         & Image & 70,000   & --      & \textcolor{red}{\xmark} & \textcolor{red}{\xmark} & \textcolor{red}{\xmark} & \textcolor{red}{\xmark} & \textcolor{green!60!black}{\cmark} & \textcolor{red}{\xmark} & \textcolor{red}{\xmark} & \textcolor{red}{\xmark}  \\
AgeDB~\cite{moschoglou2017agedb}       & Image & 16,488   & 568     & \textcolor{green!60!black}{\cmark} & \textcolor{green!60!black}{\cmark} & \textcolor{red}{\xmark} & \textcolor{red}{\xmark} & \textcolor{red}{\xmark} & \textcolor{red}{\xmark} & \textcolor{red}{\xmark} & \textcolor{red}{\xmark}  \\
RFW~\cite{wang2019racial}           & Image & 40,000   & 12,000  & \textcolor{red}{\xmark} & \textcolor{red}{\xmark} & \textcolor{green!60!black}{\cmark} & \textcolor{red}{\xmark} & \textcolor{red}{\xmark} & \textcolor{red}{\xmark} & \textcolor{red}{\xmark} & \textcolor{red}{\xmark}  \\
AffectNet~\cite{mollahosseini2017affectnet} & Image & 450,000 & --     & \textcolor{red}{\xmark} & \textcolor{red}{\xmark} & \textcolor{red}{\xmark} & \textcolor{green!60!black}{\cmark} & \textcolor{green!60!black}{\cmark} & \textcolor{red}{\xmark} & \textcolor{red}{\xmark} & \textcolor{red}{\xmark}  \\
FairFace~\cite{karkkainen2021fairface} & Image & 108,501  & --      & \textcolor{green!60!black}{\cmark} & \textcolor{green!60!black}{\cmark} & \textcolor{green!60!black}{\cmark} & \textcolor{red}{\xmark} & \textcolor{red}{\xmark} & \textcolor{red}{\xmark} & \textcolor{red}{\xmark} & \textcolor{red}{\xmark}  \\
UTKFace~\cite{zhang2017age}   & Image & 23,708   & --      & \textcolor{green!60!black}{\cmark} & \textcolor{green!60!black}{\cmark} & \textcolor{green!60!black}{\cmark} & \textcolor{red}{\xmark} & \textcolor{green!60!black}{\cmark} & \textcolor{red}{\xmark} & \textcolor{red}{\xmark} & \textcolor{red}{\xmark}  \\
GazeCapture~\cite{krafka2016eye} & Image & 2,445,504 & 1,450 & \textcolor{red}{\xmark} & \textcolor{red}{\xmark} & \textcolor{red}{\xmark} & \textcolor{red}{\xmark} & \textcolor{red}{\xmark} & \textcolor{red}{\xmark} & \textcolor{green!60!black}{\cmark} & \textcolor{red}{\xmark}  \\
PubFig~\cite{pubfig} & Image & 58,797 & 200 & \textcolor{red}{\xmark} & \textcolor{red}{\xmark} & \textcolor{red}{\xmark} & \textcolor{red}{\xmark} & \textcolor{red}{\xmark} & \textcolor{red}{\xmark} & \textcolor{red}{\xmark} & \textcolor{red}{\xmark}  \\
FERET~\cite{feret} & Image & 14,126 & 1,199 & \textcolor{red}{\xmark} & \textcolor{red}{\xmark} & \textcolor{red}{\xmark} & \textcolor{red}{\xmark} & \textcolor{red}{\xmark} & \textcolor{red}{\xmark} & \textcolor{red}{\xmark} & \textcolor{red}{\xmark}  \\
VGGFace2~\cite{vggface2} & Image & 3,310,000 & 9,131 & \textcolor{red}{\xmark} & \textcolor{red}{\xmark} & \textcolor{green!60!black}{\cmark} & \textcolor{red}{\xmark} & \textcolor{green!60!black}{\cmark} & \textcolor{red}{\xmark} & \textcolor{red}{\xmark} & \textcolor{red}{\xmark}  \\
CASIA-WebFace~\cite{casia-webface}& Image & 494,414 & 10,575 & \textcolor{red}{\xmark} & \textcolor{red}{\xmark} & \textcolor{red}{\xmark} & \textcolor{red}{\xmark} & \textcolor{red}{\xmark} & \textcolor{red}{\xmark} & \textcolor{red}{\xmark} & \textcolor{red}{\xmark}  \\
\midrule
DFME~\cite{zhao2023dfme}  & Video & 7,526 clips & 671 & \textcolor{red}{\xmark} & \textcolor{red}{\xmark} & \textcolor{red}{\xmark} & \textcolor{red}{\xmark} & \textcolor{red}{\xmark} & \textcolor{green!60!black}{\cmark} & \textcolor{red}{\xmark} & \textcolor{red}{\xmark} \\
PURE~\cite{stricker2014non}  & Video & 60 videos & 10 & \textcolor{red}{\xmark} & \textcolor{red}{\xmark} & \textcolor{red}{\xmark} & \textcolor{red}{\xmark} & \textcolor{red}{\xmark} & \textcolor{red}{\xmark} & \textcolor{red}{\xmark} & \textcolor{green!60!black}{\cmark} \\
\bottomrule
\end{tabular}%
}
\vspace{-5mm}
\end{table*}

\section{Related Work}
\subsection{FDeID Methods}
FDeID has evolved substantially from early approaches to sophisticated learning-based techniques. 
Early methods employed simple image processing operations including pixelation, blurring, and masking~\cite{boyle2000effects,hudson1996techniques,letournel2015face,wei2026face}. While computationally efficient, these approaches fundamentally trade utility for privacy, degrading visual quality and destroying most facial information.
The $k$-Same family of algorithms~\cite{gross2005integrating,gross2006model,meng2014face,newton2005preserving} introduced a principled privacy formulation based on $k$-anonymity~\cite{sweeney2002k}, synthesizing faces that match at least $k$ individuals in a database. However, these methods produce visible artifacts and offer limited control over the privacy-utility trade-off.
Recent advances in generative modeling have enabled more nuanced approaches: Generative Adversarial Networks (GAN)~\cite{goodfellow2014generative}-based methods disentangle identity from attributes~\cite{maximov2020ciagan}, inpaint faces conditioned on surrounding context~\cite{hukkelaas2023deepprivacy2}, or manipulate latent representations~\cite{agarwal2021privacy}, while diffusion models~\cite{ho2020denoising} offer improved fidelity and controllability~\cite{sun2024diffam}. 
Despite this methodological diversity, existing methods share a common evaluation blind spot: utility preservation is assessed in a fragmented manner, with each work measuring different attributes on different datasets using different metrics. Our benchmark addresses this gap through systematic, multi-level utility evaluation under a unified protocol.

\subsection{Evaluation Metrics and Protocols}

Evaluation of FDeID spans three dimensions: identity suppression, utility preservation, and image quality, each of which has accumulated
a heterogeneous collection of metrics. \textit{Identity suppression} is most commonly reported as the de-identification success
rate~\cite{sharif2016accessorize,khorzooghi2023examining,wang2024invisible,liu2025projattacker,salar2025enhancing}, with continuous or ranked
variants including recognition accuracy drop~\cite{li2019identification,zhai2022a3gan}, cosine similarity~\cite{wen2022IdentityMask,zhang2024rbgan}, identity distance~\cite{cao2023achieving,wen2023divide},
Rank-$N$-$T$~\cite{yang2021towards,wang2025adv}, and AUC~\cite{meden2023face,proencca2022uu,pan2024opticaldr}, which differ
in normalization and recognizer dependence. \textit{Utility preservation} is typically tied to the attribute being evaluated,
using categorical accuracy for expression~\cite{maximov2020ciagan} and classification or regression error for age and gender~\cite{gong2023stealthy}; these protocols need annotations from different datasets (see Tab.~\ref{tab:comparison}). \textit{Image quality} draws from reference-based metrics such as PSNR and SSIM~\cite{wang2004image}, learned perceptual metrics such as LPIPS~\cite{zhang2018unreasonable}, and distributional metrics such as FID~\cite{heusel2017gans}. This proliferation of heterogeneous metrics and protocols impedes systematic comparison and reproducibility, while no existing metric provides a unified assessment across all three dimensions~\cite{wei2026fdeid}. \metric\ addresses both by adopting a consistency-based formulation and aggregating all three dimensions into a single interpretable score.

\subsection{Facial Signal Analysis}
Facial images encode abundant information spanning identity features and non-identity utility cues across multiple granularities. 
At the coarsest level, well-established tasks such as age estimation~\cite{geng2007automatic}, gender classification~\cite{makinen2008evaluation}, ethnicity recognition~\cite{karkkainen2021fairface}, expression recognition~\cite{li2020deep}, and facial landmark detection~\cite{li2022towards} represent mature fields with established benchmarks and high-performing models; these macro-level signals are spatially coarse, semantically salient, and readily perceivable by human observers. 
At a finer level, micro-expressions~\cite{zhao2023dfme} and gaze direction~\cite{cheng2024appearance} capture subtle behavioral and attentional states that are critical for affective computing, deception detection, and human-computer interaction; these signals depend on precise localized geometry and temporal dynamics that are far more fragile under facial manipulation. At the finest level, deep learning approaches for rPPG, including DeepPhys~\cite{chen2018deepphys}, PhysNet~\cite{yu2019remote}, and EfficientPhys~\cite{liu2023efficientphys}, now recover heart rate and physiological waveforms from sub-pixel skin color variations in face videos, with transformative potential for telehealth and remote patient monitoring. 

This natural stratification, from coarse attributes perceivable at a glance, through subtle behavioral dynamics, to physiological signals invisible to the naked eye, suggests an inherent hierarchy of facial signal that FDeID methods affect at different rates. Yet this hierarchical structure remains unexplored in FDeID evaluation. We formalize this observation and introduce a three-level utility framework that enables principled assessment of what facial information FDeID methods preserve or destroy.

\section{Proposed Protocol}

\subsection{\uface{} Benchmark}
\label{subsec:construction}
A benchmark suitable for comprehensive FDeID evaluation must simultaneously provide sufficient identity diversity for meaningful
privacy evaluation, adequate image quality for reliable attribute estimation, and demographic balance to avoid evaluation bias.
No existing face dataset satisfies these three requirements jointly (Tab.~\ref{tab:comparison}).
We therefore design a four-stage curation pipeline that aggregates, cleans, enhances, and balances face images from four large-scale
sources; a summary of each stage and its output statistics is reported in Tab.~\ref{tab:pipeline}, and full operational details
(restoration-tool selection, identity-deduplication threshold, quality-filter justification) are deferred to Appendix~\ref{sec:app_benchmark}.

\noindent\textbf{Source aggregation.}
We aggregate WebFace42M~\cite{zhu2021webface260m}, Glint360K~\cite{an2021partial}, MS1MV2~\cite{deng2019arcface}, and
MegaFace~\cite{kemelmacher2016megaface}, yielding an unfiltered pool of approximately $69.55$M images across $3.12$M identities.
These datasets were originally curated to train robust face recognition models and thus contain substantial proportions of challenging samples
(heavy blur, extreme pose, occlusion) that, while beneficial for recognition training, are unsuitable for utility-aware FDeID evaluation.

\noindent\textbf{Identity-aware cleaning and resolution enhancement.}
We discard samples that would compromise downstream evaluation: images with zero or multiple detected faces, non-RGB inputs, heavily blurred
samples, and cross-dataset duplicate identities identified via pretrained ArcFace~\cite{deng2019arcface} similarity.
Surviving low-resolution samples are upsampled to $448{\times}448$ or $512{\times}512$ using blind face
restoration~\cite{wang2021towards,zhou2022towards}, with the upsampling ratio capped at $4{\times}$ to preserve perceptual fidelity.
This stage yields $\sim 20.58$M images across $\sim 701{,}810$ identities.

\noindent\textbf{Perceptual quality filtering.}
We apply a quality filter based on NIQE~\cite{mittal2012making}, discarding images with $\text{NIQE} \geq 5.0$,
an empirically motivated cut-off that admits naturalistic faces while rejecting restoration failures.
The pool contains $\sim 8.69$M images across $\sim 640{,}035$ identities after this stage.

\begin{figure}[t]
    \centering
    \includegraphics[width=1.0\linewidth]{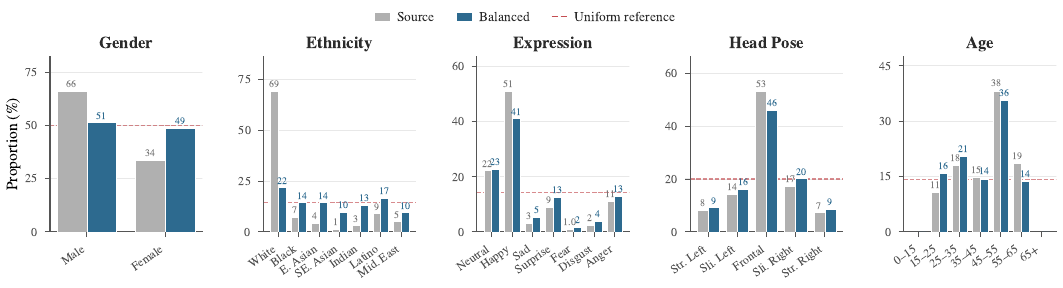}
    \vspace{-6mm}
    \caption{\textbf{Distributions comparison before and after the balance sampling.} Each panel reports the proportion of images across categories for five facial annotations: gender, ethnicity, expression, head pose, and age. }
    \label{fig:statistics}
    \vspace{-5mm}
\end{figure}

\noindent\textbf{Demographically balanced benchmark construction.}
The remaining pool exhibits strong distributional skew across gender, ethnicity, expression, pose, and age (Fig.~\ref{fig:statistics}, grey
bars), which would otherwise conflate method performance with demographic bias.
We apply a hierarchical stratified sampling strategy operating at the identity level: identities are partitioned into $14$ strata defined by
gender and ethnicity, then ranked within each stratum by a composite score combining image count and diversity over
expression and head pose.
The top-ranked identities fill an equal per-stratum budget, and all images of each selected identity are retained.
The resulting \uface\ benchmark contains $99{,}928$ images across $2{,}069$ identities (mean $48.3$, median $30$ images per identity),
with near-uniform gender balance ($51.5\%$ / $48.5\%$) and a flattened ethnicity distribution that reduces the largest-to-smallest category
ratio from $53\times$ to $2.3\times$ (Fig.~\ref{fig:statistics}, teal bars).

Unlike the image datasets in Tab.~\ref{tab:comparison}, \uface\ carries no attribute annotations by design: the consistency-based \metric\
metric measures the consistency between estimator outputs on $\mathbf{x}$ and $\hat{\mathbf{x}}$, so ground-truth labels
are neither needed nor desirable.

\begin{table}[t]
\centering
\caption{\textbf{\uface\ curation pipeline.} Each stage progressively filters the pool toward a high-quality, demographically balanced evaluation benchmark.}
\label{tab:pipeline}
\resizebox{0.95\columnwidth}{!}{%
\begin{tabular}{@{}llrr@{}}
\toprule
\textbf{Stage} & \textbf{Operation} & \textbf{\#Images} & \textbf{\#Identities} \\
\midrule
1. Aggregation                & Merge 4 source datasets                       & $\sim 69.55$M & $\sim 3.12$M \\
2. Cleaning + Enhancement     & Identity-aware filtering; GFPGAN / CodeFormer & $\sim 20.58$M & $\sim 701{,}810$ \\
3. Quality filter             & NIQE $< 5.0$                                  & $\sim 8.69$M  & $\sim 640{,}035$ \\
4. Balanced sampling          & Hierarchical stratified sampling              & $99{,}928$    & $2{,}069$ \\
\bottomrule
\end{tabular}%
}
\vspace{-5mm}
\end{table}

\subsection{\metric{} Metric}
\label{sec:fdeidscore}

Evaluating FDeID requires jointly quantifying two competing objectives: \emph{how effectively identity is suppressed} and \emph{how faithfully
non-identity information is retained}.
Existing protocols report these objectives in isolation, treat utility as a monolithic concept, and often rely on ground-truth
labels.
We address these limitations with \metric, a unified scalar metric that integrates identity suppression, hierarchical utility
preservation, and image quality into a single interpretable score.
\metric\ adopts a \emph{consistency-based} formulation throughout: every component is computed as the consistency between a pretrained estimator's outputs on the original input and its de-identified counterpart, directly quantifying the amount of downstream-perceivable information that survives de-identification.
We first define a generic consistency function, instantiate it on the three-level hierarchy, then specify privacy and quality, and combine all five into the composite \metric\ score.

\noindent\textbf{Consistency design principle.}
For the $k$-th estimator with output space $\mathcal{Y}_k$, a \emph{consistency function} $s_k: \mathcal{Y}_k \times \mathcal{Y}_k \rightarrow [0, 1]$ takes the estimator's outputs on the original face $\mathbf{x}$ and its de-identified counterpart $\hat{\mathbf{x}}$ and returns their agreement as a similarity score, with $1$ indicating identical outputs.
We require every $s_k$ to satisfy:
(i)~\emph{boundedness}, $s_k \in [0,1]$ with $s_k = 1$ when the outputs are identical;
(ii)~\emph{type-appropriateness}, tailored to the output modality (scalar, categorical distribution, coordinate array, or temporal signal);
(iii)~\emph{graceful degradation}, with a well-defined value at degenerate inputs.
The metric does not require per-image ground-truth labels; the only quantities specified a priori are global normalization constants
(attribute ranges, error caps) that are hand-defined once for each task, based on domain conventions.
See details in Appendix~\ref{sec:app_metric_agreements}.

\noindent\textbf{Hierarchical utility scores.}
We instantiate the principle above on the three-level hierarchy of Fig.~\ref{fig:hierarchy}, yielding five L1, two L2, and two L3 consistency scores.
\textbf{L1 (macro-cues)} covers age, gender, ethnicity, macro-expression, and landmark consistency, using truncated absolute error for scalars, total variation distance for categorical posterior distributions (with a top-1-plus-confidence binary posterior reconstructed for gender), and normalized mean error (NME) for landmark coordinates anchored on the inter-ocular distance of $\mathbf{x}$; we aggregate into $U_1$ by unweighted averaging.
\textbf{L2 (micro-cues)} covers gaze and micro-expression consistency, aggregated into $U_2$ as the mean when video is available and as
$s_{\text{gaze}}$ alone for image-only inputs.
\textbf{L3 (imperceptible cues)} covers blood-volume-pulse (BVP) and heart-rate (HR) consistency from rPPG signals, aggregated into $U_3$
after averaging over an ensemble of rPPG estimators
(PhysNet~\cite{yu2019remote},
EfficientPhys~\cite{liu2023efficientphys},
FactorizePhys~\cite{joshi2024factorizephys}).
BVP uses a positive-correlation rescaling that assigns zero consistency to uncorrelated or anti-correlated pairs, and HR uses a tightened cap of $\tau_{\text{HR}} = 20$\,bpm to match the clinical relevance threshold.

\noindent\textbf{Privacy and quality.}
$\bar{P}$ measures identity suppression as the mean embedding-space
dissimilarity across an ensemble of $M$ face recognizers (ArcFace~\cite{deng2019arcface}, CosFace~\cite{wang2018cosface}, AdaFace~\cite{kim2022adaface}), rescaled from cosine similarity to $[0, 1]$; the ensemble mitigates the known vulnerability of single-recognizer evaluation against adversarial methods.
$Q$ combines a reference-based perceptual similarity (LPIPS~\cite{zhang2018unreasonable}) and a no-reference naturalness
score (NIQE~\cite{mittal2012making}) with equal weight, both bounded to $[0, 1]$.

\noindent\textbf{Composite \metric\ score.}
All five dimensions ($\bar{P}$, $Q$, $U_1, U_2, U_3$) are bounded in $[0, 1]$ by construction, allowing a weighted harmonic mean
aggregation that penalizes imbalanced methods more strongly than arithmetic or geometric averaging:
\begin{equation}
    \text{\metric} = \biggl(w_P + w_Q + \sum_{\ell=1}^{L} w_\ell\biggr) \cdot \biggl(\frac{w_P}{\bar{P} + \epsilon} + \frac{w_Q}{Q + \epsilon} + \sum_{\ell=1}^{L} \frac{w_\ell}{U_\ell + \epsilon}\biggr)^{-1},
    \label{eq:fdeid}
\end{equation}
where $\mathbf{w} = (w_P, w_Q, w_1, \dots, w_L)$ are non-negative importance weights, $\epsilon = 10^{-6}$ prevents division by zero, and 
$L \in \{1, 2, 3\}$ is the number of hierarchy levels scoreable for the given input: $L=1$ for images without gaze, $L=2$ for images with gaze or video without rPPG support, and $L=3$ for video supporting rPPG.
The outer factor $(\sum w)$ renders the score self-normalizing. 
We define three weighting profiles:
(i)~\textit{Privacy-First} ($w_P{=}0.5$, remainder equal) for high-risk surveillance;
(ii)~\textit{Balanced} (uniform weights), used as the default;
(iii)~\textit{Clinical} ($w_3{=}0.4$, remainder equal) for telehealth applications prioritizing physiological-signal preservation.

\noindent\textbf{Properties.}
By construction, \metric\ satisfies:
(i)~\emph{boundedness}, $\text{\metric} \in [0, 1]$;
(ii)~\emph{monotonicity}, improving any component increases the score;
(iii)~\emph{vetoing}, the harmonic mean drives the score toward zero when any single component approaches zero;
(iv)~\emph{graceful degradation} across modalities via $L$;
(v)~\emph{annotation-free}, every component depends only on estimator outputs on $\mathbf{x}$ and $\hat{\mathbf{x}}$.

\section{Comparative Analysis}
\subsection{Experimental Setup}

\noindent\textbf{FDeID methods.}
We evaluate 12 representative methods spanning the three dominant FDeID paradigms based on their underlying generation mechanism.
(i)~\emph{Adversarial perturbation}, which applies $\varepsilon$-bounded gradient-based noise to fool face recognition models: MI-FGSM~\cite{dong2018boosting}, PGD~\cite{madry2018towards}, TI-DIM~\cite{dong2019evading}, TIP-IM~\cite{yang2021towards}, and Chameleon~\cite{chow2024personalized}.
(ii)~\emph{GAN-based generation}, which produces anonymized images via a learned generator trained with adversarial objectives: CIAGAN~\cite{maximov2020ciagan} for identity-conditioned replacement, AMT-GAN~\cite{hu2022protecting} for adversarial makeup transfer, Adv-Makeup~\cite{yin2021adv} for eye-region makeup, DeID-rPPG~\cite{savic2023identification} for heart-rate preservation, and G$^{2}$Face~\cite{yang2024g} for geometry and expression preservation.
(iii)~\emph{Diffusion-based generation}, represented by DiffAM~\cite{sun2024diffam} and WeakenDiff~\cite{salar2025enhancing}, anonymizes faces through iterative denoising sampling.

\begin{table}[t]
\centering
\caption{\textbf{\metric\ comparative results.} Consistency scores (higher is better, $\in[0,1]$) of each method under the \metric\ protocol. Sub-scores are reported per dimension; $U_1$ averages the five L1 consistency scores, $U_2$ averages gaze and micro-expression, and $U_3$ averages BVP and HR. The final column is the \metric\ composite.} 
\label{tab:hifd_main}
\resizebox{\textwidth}{!}{%
\begin{tabular}{@{}l cc ccccc c cc c cc c c@{}}
\toprule
\multirow{2}{*}{\textbf{Method}} & \multirow{2}{*}{$\bar{P}\!\uparrow$} & \multirow{2}{*}{$Q\!\uparrow$}
  & \multicolumn{5}{c}{\textbf{L1: Macro-cues}} & \multirow{2}{*}{$U_1\!\uparrow$}
  & \multicolumn{2}{c}{\textbf{L2: Micro-cues}} & \multirow{2}{*}{$U_2\!\uparrow$}
  & \multicolumn{2}{c}{\textbf{L3: rPPG}} & \multirow{2}{*}{$U_3\!\uparrow$}
  & \multirow{2}{*}{\metric$\!\uparrow$} \\
\cmidrule(lr){4-8}\cmidrule(lr){10-11}\cmidrule(lr){13-14}
 & & & Age & Gen. & Eth. & MExp & Lmk & & Gaze & ME & & BVP & HR & & \\
\midrule
\multicolumn{16}{l}{\textit{(a) Adversarial perturbation}} \\
MI-FGSM       & 0.134 & 0.360 & 0.975 & 0.962 & 0.916 & 0.951 & 0.834 & 0.928 & 0.936 & 0.799 & 0.868 & 0.471 & 0.478 & 0.474 & 0.343\\
PGD           & 0.113 & 0.362 & 0.978 & 0.968 & 0.935 & 0.964 & 0.843 & 0.938 & 0.940 & 0.817 & 0.878 & 0.577 & 0.555 & 0.566 & 0.321\\
TI-DIM        & \textbf{0.500} & 0.342 & 0.880 & 0.826 & 0.656 & 0.739 & 0.684 & 0.757 & 0.921 & 0.722 & 0.821 & 0.424 & 0.425 & 0.424 & 0.509\\
TIP-IM        & 0.496 & 0.446 & 0.927 & 0.882 & 0.701 & 0.794 & 0.728 & 0.806 & 0.944 & 0.812 & 0.878 & 0.643 & 0.605 & 0.624 & \textbf{0.607}\\
Chameleon     & 0.239 & 0.474 & 0.915 & 0.897 & 0.780 & 0.839 & 0.816 & 0.850 & 0.961 & 0.816 & 0.889 & 0.821 & 0.777 & 0.799 & 0.508\\
\midrule
\multicolumn{16}{l}{\textit{(b) GAN-based generation}} \\
Adv-Makeup    & 0.012 & \textbf{0.524} & \textbf{0.987} & \textbf{0.980} & \textbf{0.942} & \textbf{0.971} & \textbf{0.917} & \textbf{0.960} & \textbf{0.966} & \textbf{0.916} & \textbf{0.941} & \textbf{0.964} & \textbf{0.898} & \textbf{0.931} & 0.056\\
AMT-GAN       & 0.156 & 0.481 & 0.966 & 0.911 & 0.737 & 0.868 & 0.821 & 0.861 & 0.938 & 0.553 & 0.745 & 0.132 & 0.164 & 0.148 & 0.282\\
CIAGAN        & 0.450 & 0.276 & 0.894 & 0.549 & 0.291 & 0.529 & 0.075 & 0.468 & 0.748 & 0.572 & 0.660 & 0.225 & 0.269 & 0.247 & 0.369\\
DeID-rPPG     & 0.013 & 0.407 & 0.980 & 0.971 & 0.890 & 0.963 & 0.854 & 0.932 & 0.940 & 0.838 & 0.889 & 0.940 & 0.875 & 0.907 & 0.061\\
G$^{2}$Face   & 0.210 & 0.443 & 0.953 & 0.898 & 0.712 & 0.817 & 0.665 & 0.809 & 0.903 & 0.793 & 0.848 & 0.324 & 0.328 & 0.326 & 0.400\\
\midrule
\multicolumn{16}{l}{\textit{(c) Diffusion-based generation}} \\
WeakenDiff    & 0.222 & 0.371 & 0.951 & 0.922 & 0.680 & 0.865 & 0.754 & 0.834 & 0.924 & 0.659 & 0.792 & 0.169 & 0.201 & 0.185 & 0.332\\
DiffAM        & 0.159 & 0.471 & 0.959 & 0.923 & 0.698 & 0.802 & 0.583 & 0.793 & 0.908 & 0.715 & 0.811 & 0.538 & 0.575 & 0.557 & 0.394\\
\bottomrule
\end{tabular}
}
\vspace{-6mm}
\end{table}

\noindent\textbf{Datasets.}
All privacy, L1 attributes, L2 gaze, and quality metrics are computed on \uface\ ($99{,}928$ images, $2{,}069$ identities).
L2 micro-expression is evaluated on DFME~\cite{zhao2023dfme} test-set A ($474$ video clips).
L3 rPPG is evaluated on PURE~\cite{stricker2014non} ($10$ subjects, $60$ video clips).
Additionally, we use RAF-DB~\cite{li2017reliable} and MMPD~\cite{tang2023mmpd} for the \metric\ validation.

\noindent\textbf{Estimators.}
Every score in \metric\ is a consistency score between a pretrained estimator's outputs on the original and the de-identified input.
\emph{Privacy} uses an ensemble of ArcFace~\cite{deng2019arcface}, CosFace~\cite{wang2018cosface}, and AdaFace~\cite{kim2022adaface} face recognizers, with face bounding boxes and 5-point landmarks cached once on the original via RetinaFace~\cite{deng2020retinaface} and reused on the de-identified counterparts for consistent alignment.
\emph{L1}: age and gender via MiVOLO~\cite{kuprashevich2023mivolo} (trained on UTKFace), ethnicity via FairFace~\cite{karkkainen2021fairface}, macro-expression via POSTER~\cite{zheng2023poster} (trained on RAF-DB), and 68-point facial landmarks via MediaPipe~\cite{lugaresi2019mediapipe} Face Landmarker.
\emph{L2}: gaze estimation via L2CS-Net~\cite{abdelrahman2023l2cs} (trained on Gaze360~\cite{kellnhofer2019gaze360}), and micro-expression via SAMER~\cite{ruan2022mimicking} (five-class).
\emph{L3}: BVP waveforms via PhysNet~\cite{yu2019remote}, EfficientPhys~\cite{liu2023efficientphys}, and FactorizePhys~\cite{joshi2024factorizephys} (all trained on UBFC-rPPG dataset).
\emph{Quality}: LPIPS-VGG~\cite{zhang2018unreasonable} and NIQE~\cite{mittal2012making} computed via \texttt{pyiqa}.

\noindent\textbf{Implementation details.}
All inference and scoring runs on 8 AMD MI250X GPUs.
\metric\ components are aggregated per method (privacy and per-image sub-scores as mean over $99{,}928$ keys; rPPG as mean over 60 clips $\times$ $3$ estimators; micro-expression as mean over 474 clips) and combined via the weighted harmonic mean in Eq.~\eqref{eq:fdeid}.
Unless otherwise noted, we report the Balanced profile ($w_P = w_Q = w_1 = w_2 = w_3$).
Normalization caps follow §\ref{sec:fdeidscore}: $\Delta_{\text{age}} = 100$, $\tau_{\text{NME}} = 0.10$, $\theta_{\max} = 90^\circ$, $\tau_{\text{HR}} = 20$\,bpm, $\tau_{\text{N}} = 5$, $\alpha = 0.5$.
See Appendix for more details.

\vspace{-2mm}
\subsection{Main Results}
Tab.~\ref{tab:hifd_main} reports \metric\ and all sub-scores across the 12 evaluated methods. Five findings emerge that are invisible under existing fragmented protocols.

\noindent\textbf{Finding 1: The privacy-utility frontier is not monotonic across paradigms.}
No FDeID family dominates on both axes and each trades privacy for utility on its own terms.
Adversarial methods retain the highest macro-cue scores ($U_1 > 0.92$ for PGD and MI-FGSM), as expected from their minimally visible pixel perturbations, but only the two with explicit transferability objectives reach $\bar{P} \approx 0.50$ (TI-DIM, TIP-IM). MI-FGSM and PGD stall at $\bar{P} < 0.14$, suggesting that recent robust recognizers largely survive naive gradient attacks; in operational terms they de-identify only $8\%$ of faces at any false-accept rate (Appendix~\ref{sec:app_far}).
GAN methods cover the widest privacy range: CIAGAN reaches $\bar{P} = 0.45$ by replacing the face entirely, but at severe cost on $U_1$ ($0.47$) and landmarks ($0.08$), while cosmetic methods such as Adv-Makeup preserve macro-cues almost perfectly yet offer virtually no identity suppression ($\bar{P} = 0.01$).
Diffusion methods sit between these extremes. WeakenDiff achieves $\bar{P} = 0.22$ and $U_1 = 0.83$. DiffAM, despite its explicit adversarial-makeup objective, scores only $\bar{P} = 0.16$ on our ArcFace/CosFace/AdaFace ensemble, since its makeup perturbations were trained against IRSE50/IR152/MobileFace/FaceNet and transfer imperfectly to modern recognizers. DiffAM nonetheless outperforms WeakenDiff on $U_2$, $U_3$, and $Q$ ($0.81$, $0.56$, $0.47$) and trails only marginally on $U_1$ ($0.79$ vs.\ $0.83$), yielding a higher \metric\ of $0.394$ versus $0.332$.

\begin{figure}
\begin{minipage}[t]{0.65\linewidth}
    \centering
    \includegraphics[width=\linewidth]{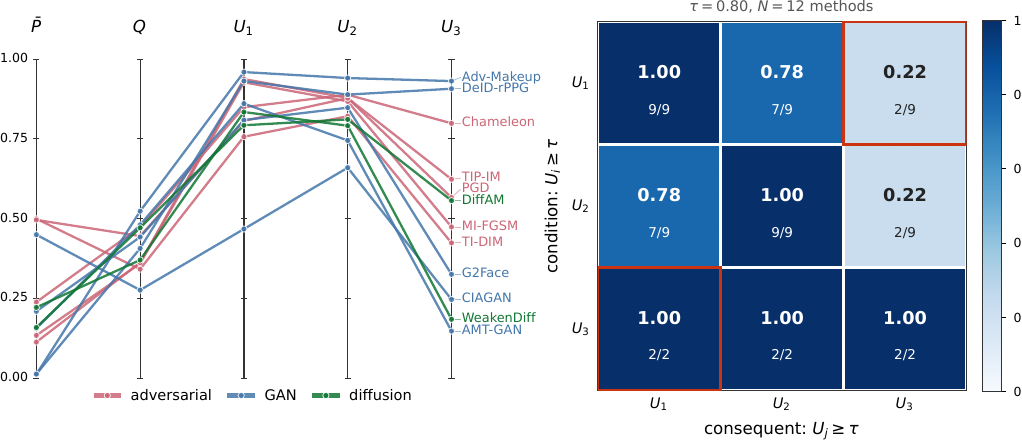}
    \vspace{-5mm}
    \caption{\textbf{Hierarchical utility preservation is asymmetrically coupled.}
    \textbf{(Left)} Parallel coordinates plot over the five \metric\ axes ($\bar{P}$, $Q$, $U_1$, $U_2$, $U_3$).
    \textbf{(Right)} Preservation matrix at threshold $\tau\!=\!0.80$. Cell $(i,j)$ reports $\Pr(U_j\!\geq\!\tau\,|\,U_i\!\geq\!\tau)$ across the 12 methods, with the intersection over marginal count shown. Red boxes highlight the asymmetric pair $(U_1, U_3)$: $\Pr(U_1\!\geq\!\tau\,|\,U_3\!\geq\!\tau){=}1.00$ but $\Pr(U_3\!\geq\!\tau\,|\,U_1\!\geq\!\tau){=}0.22$.}
    \label{fig:utility_correlation}
\end{minipage} %
\hspace{1mm}
\begin{minipage}[b]{0.34\linewidth}
  \centering
  \captionsetup{type=table}
  \caption{\textbf{\metric\ under the three profiles}. Privacy-First: $w_P{=}0.5$; Balanced: uniform; Clinical: $w_3{=}0.4$; remaining weights distributed equally.}
  \vspace{-2mm}
  \setlength{\tabcolsep}{5pt}
  \resizebox{\textwidth}{!}{
  \label{tab:hifd_profiles}
    \begin{tabular}{@{}lccc@{}}
    \toprule
    \textbf{Method} & \textbf{Privacy-First} & \textbf{Balanced} & \textbf{Clinical} \\
    \midrule
    \multicolumn{4}{l}{\textit{(a) Adversarial perturbation}} \\
    MI-FGSM       & 0.217 & 0.343 & 0.369 \\
    PGD           & 0.190 & 0.321 & 0.360 \\
    TI-DIM        & 0.506 & 0.509 & 0.485 \\
    TIP-IM        & \textbf{0.560} & \textbf{0.607} & \textbf{0.611} \\
    Chameleon     & 0.357 & 0.508 & 0.558 \\
    \midrule
    \multicolumn{4}{l}{\textit{(b) GAN-based generation}} \\
    Adv-Makeup    & 0.023 & 0.056 & 0.073 \\
    AMT-GAN       & 0.217 & 0.282 & 0.230 \\
    CIAGAN        & 0.396 & 0.369 & 0.328 \\
    DeID-rPPG     & 0.026 & 0.061 & 0.079 \\
    G$^{2}$Face   & 0.298 & 0.400 & 0.378 \\
    \midrule
    \multicolumn{4}{l}{\textit{(c) Diffusion-based generation}} \\
    WeakenDiff    & 0.280 & 0.332 & 0.277 \\
    DiffAM        & 0.254 & 0.394 & 0.425 \\
    \bottomrule
    \end{tabular}
  }
\end{minipage}%
\vspace{-7mm}
\end{figure}

\noindent\textbf{Finding 2: Hierarchical signals degrade at different rates.}
Preservation drops from L1 to L3 for almost every method, with the steepest collapse in generative families.
AMT-GAN, CIAGAN, G$^{2}$Face, and WeakenDiff all score $U_3 < 0.33$, confirming that face synthesis decouples macro appearance from fine-grained skin-color dynamics.
Adversarial methods preserve $U_3$ best (TIP-IM $0.62$, Chameleon $0.80$), since bounded $\ell_\infty$ or $\ell_2$ noise leaves low-frequency skin signals largely intact.
Only Adv-Makeup ($U_3\!=\!0.93$) and DeID-rPPG ($U_3\!=\!0.91$) combine high $U_3$ with high $U_1$, but both trade this preservation for a complete loss of identity suppression.

\noindent\textbf{Finding 3: Hierarchical preservation is asymmetrically coupled.}
\textit{Does retaining the strictest level $U_3$ imply retention of $U_2$ and $U_1$, and conversely?}
Fig.~\ref{fig:utility_correlation} answers across the 12 methods.
The parallel-coordinates panel (left) shows the polylines fanning out from $U_1$ to $U_3$: nine of the twelve methods exceed $0.80$ on $U_1$ and nine on $U_2$, yet only Adv-Makeup and DeID-rPPG sit above $U_3\!=\!0.90$, while AMT-GAN, CIAGAN, WeakenDiff, and G$^{2}$Face fall below $U_3\!=\!0.35$.
The conditional preservation matrix (right) quantifies the implication structure at threshold $\tau\!=\!0.80$: every method that preserves $U_3$ also preserves $U_1$ and $U_2$ ($\Pr(U_1\!\geq\!\tau\,|\,U_3\!\geq\!\tau)=\Pr(U_2\!\geq\!\tau\,|\,U_3\!\geq\!\tau)=1.00$), yet only $22\%$ (two of nine) of the methods that preserve $U_1$ also preserve $U_3$; the asymmetry holds for every $\tau\in[0.70,0.90]$. 
Practically, the asymmetry refutes the common shortcut of certifying L3 preservation by inspecting macro cues alone. This is precisely what the harmonic-mean aggregation enforces via the vetoing property: a near-zero $U_3$ component cannot be masked by strong $U_1$ scores, so methods are ranked by their weakest dimension rather than their average. TIP-IM achieves the highest \metric\ of $0.607$ for this reason: it is the only method that reaches at least $0.44$ on every dimension simultaneously, while specialized methods such as DeID-rPPG ($U_3\!=\!0.91$) and G$^{2}$Face ($U_1\!=\!0.81$) excel on a single axis but do not dominate the composite.

\noindent\textbf{Finding 4: Application-specific profiles reshape the ranking.}
Tab.~\ref{tab:hifd_profiles} reports \metric\ under the three profiles defined in §\ref{sec:fdeidscore}.
Under \emph{Privacy-First} ($w_P\!=\!0.5$), CIAGAN rises within the GAN family thanks to its strong identity suppression, yet TIP-IM still leads overall because its balanced profile avoids any near-zero component.
Under \emph{Clinical} ($w_3\!=\!0.4$), adversarial methods (TIP-IM, Chameleon) dominate as they retain physiological rPPG signals, while most GAN-based methods collapse since their face synthesis destroys sub-pixel dynamics regardless of visual fidelity.
Adv-Makeup and DeID-rPPG remain at the bottom across all profiles: despite excellent utility preservation, their near-zero identity suppression ($\bar{P} < 0.02$) is mathematically penalized by the harmonic-mean aggregation.
The optimal FDeID choice therefore depends on which utility levels the downstream application prioritizes, but no profile rescues a method that fails on identity suppression.

\begin{figure}[t]
    \centering
    \includegraphics[width=1.0\linewidth]{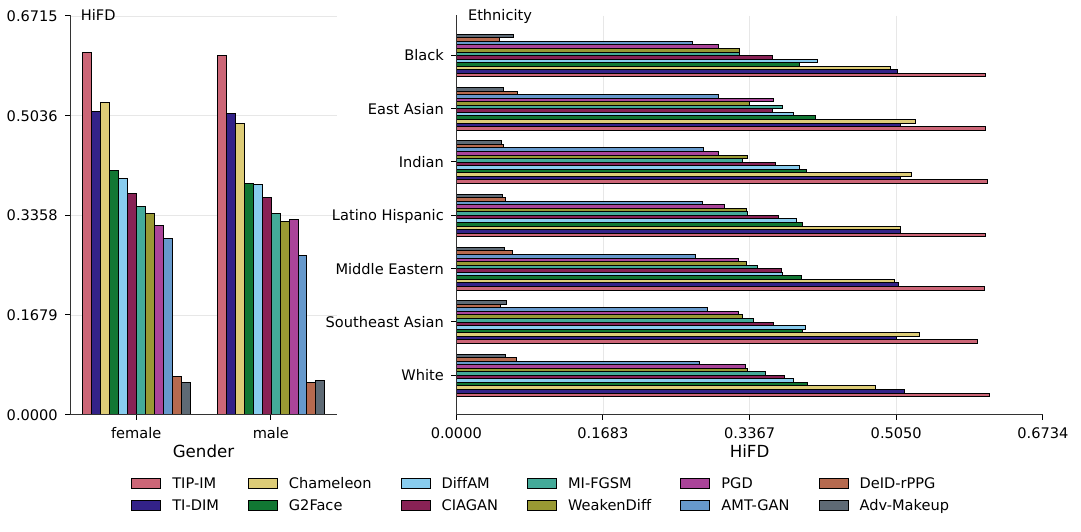}
    \vspace{-7mm}
    \caption{\textbf{Per-demographic-slice \metric\ on \uface.} Methods are ordered by global \metric\ (descending). Left shows the female/male split and right shows the seven-way ethnicity split (alphabetical). Differences across demographic strata are small in absolute terms ($<\!0.07$ within any method) compared to the inter-method spread.}
    \label{fig:hifd_demographic}
    \vspace{-5mm}
\end{figure}

\noindent\textbf{Finding 5: FDeID methods are largely demographic-agnostic, with adversarial methods showing the largest disparities.}
Fig.~\ref{fig:hifd_demographic} decomposes each method's \metric\ score across two genders and seven ethnicities.
Across all 12 methods, no single method shows a per-gender gap larger than $0.035$ or a seven-way ethnicity range larger than $0.064$, both an order of magnitude below the inter-method spread in Tab.~\ref{tab:hifd_profiles}.
The four methods with the widest per-method ethnicity range, PGD ($0.064$), Chameleon ($0.050$), MI-FGSM ($0.049$), and DiffAM ($0.039$), are all gradient-based or makeup-style perturbations that depend on local skin-tone gradients, which plausibly explains why they treat East Asian samples slightly differently from the rest. 
The top-ranked TIP-IM (gender gap $0.006$, ethnicity range $0.014$) and the bottom-ranked Adv-Makeup and DeID-rPPG
(disparities $\leq\!0.021$) confirm that neither leadership nor failure is concentrated on any subgroup.
The results demonstrate that current FDeID methods, despite spanning three different generation paradigms, do not exhibit the demographic-specific failure modes that have been documented for face recognition itself.

\subsection{\metric\ Validation}
\label{sec:validity}

\begin{figure}[t]
    \centering
    \includegraphics[width=\linewidth]{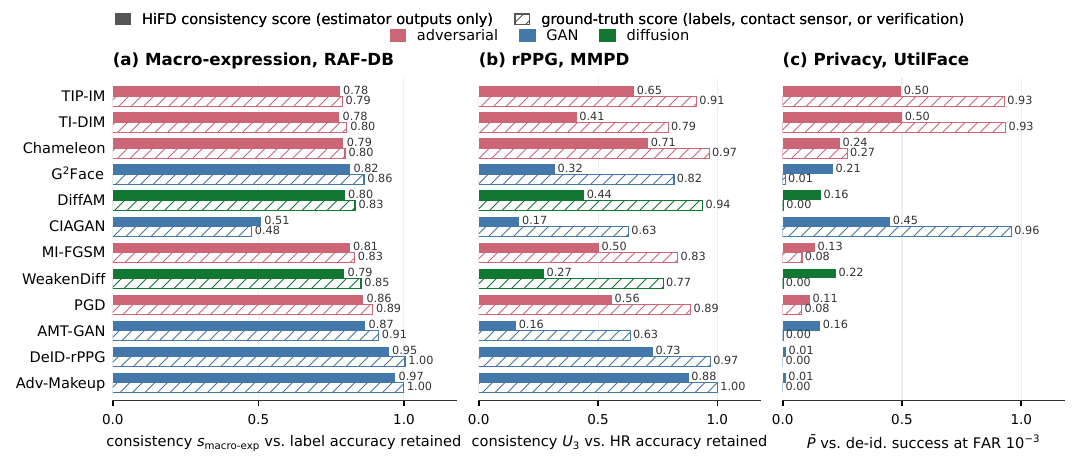}
    \vspace{-7mm}
    \caption{\textbf{Consistency scores next to ground-truth scores.} \textbf{(a)}~Macro-expression consistency on the de-identified RAF-DB against the fraction of ground-truth label accuracy retained. \textbf{(b)}~$U_3$ on MMPD against the fraction of contact-sensor heart-rate accuracy retained. \textbf{(c)}~Privacy score $\bar{P}$ against FDeID success at FAR $10^{-3}$.}
    \label{fig:validity}
    \vspace{-4mm}
\end{figure}

Every \metric\ component is a \emph{consistency score}, computed from a pretrained estimator's outputs on $\mathbf{x}$ and $\hat{\mathbf{x}}$ without any label. Its validity therefore rests on two questions: does the consistency score track \emph{true} utility preservation, and does the privacy score correspond to an operational notion of privacy? Fig.~\ref{fig:validity} places the consistency score next to a ground-truth score for every method, using ground truth that is available outside \uface. For macro-expression (Fig.~\ref{fig:validity}a) we de-identified the labeled RAF-DB test set ($3{,}068$ images) with all twelve methods and compared the consistency score with the fraction of ground-truth accuracy the same estimator retains on the de-identified images: the two bars differ by at most $0.06$ for every method, both place CIAGAN far below the rest, and both agree on the three best-preserving methods. Per image, the consistency score predicts whether the true label survives with AUC $0.98$ over $28{,}614$ pairs, and only $1.1\%$ of pairs with a consistency score above $0.8$ hide a flipped label. For rPPG (Fig.~\ref{fig:validity}b), MMPD's contact sensor gives the true heart rate, so the ground-truth bar is the fraction of heart-rate accuracy retained after de-identification. 
The two scores live on different scales, because heart-rate error saturates, but they order the methods the same way: the same three best (Adv-Makeup, DeID-rPPG, Chameleon) and the same three worst (AMT-GAN, CIAGAN, WeakenDiff), with eleven of twelve methods within one rank of each other.
A consistency score can still mislead for near-tie posteriors, which flip the hard label under an arbitrarily small total-variation change, or for an estimator that is degenerate on the de-identified domain. The first is why we measured hidden flips, and the second is bounded by estimator substitution (Appendix~\ref{sec:app_estsub}).

For privacy (Fig.~\ref{fig:validity}c) we fit verification thresholds on 3M different-identity impostor pairs and measure de-identification success, the fraction of faces no longer verified as their original. The bars show what $\bar{P}$ means in practice: the three methods with $\bar{P} \ge 0.45$ (TI-DIM, TIP-IM, CIAGAN) de-identify $93$ to $96\%$ of faces at FAR $10^{-3}$, the seven methods with $\bar{P}$ between $0.11$ and $0.24$ de-identify at most $27\%$ (three of them under $1\%$), and the two with $\bar{P} \approx 0.01$ de-identify none. Within the middle band the mapping is not monotone (MI-FGSM and PGD reach $8\%$ at $\bar{P} \approx 0.12$ while WeakenDiff stays below $1\%$ at $0.22$), which is what a continuous embedding margin produces when it meets a thresholded decision. The practitioner-level reading is nevertheless unambiguous: only three of twelve methods de-identify usefully. 
Extended evaluations are detailed in Appendix~\ref{sec:app_robustness} and \ref{sec:app_human}.

\subsection{How Private is Private?}

Our comparative study delivers a sobering answer: under the unified \metric\ protocol, no current FDeID method is simultaneously private and utility-preserving across all three hierarchical levels. The best-performing method in our study, TIP-IM, attains a \metric\ of only $0.607$, meaning that even the strongest baseline leaves a measurable gap on at least one of the five axes ($\bar{P}, Q, U_1, U_2, U_3$); every other method we evaluate scores below $0.52$. The privacy axis itself is the most fragile: nine of the twelve methods achieve $\bar{P} < 0.30$ against our three-recognizer ensemble, and only three de-identify more than $80\%$ of faces at a false-accept rate of $10^{-2}$ (Tab.~\ref{tab:app_far}), indicating that perceived ``privacy'' in much of the prior literature reflects evaluation against a single, often outdated recognizer rather than genuine identity suppression against modern systems. When privacy is achieved, it is typically purchased at a steep and uneven cost across the hierarchy, with macro cues (L1) most often retained, micro cues (L2) partially degraded, and physiological signals (L3) silently destroyed by virtually all generative approaches regardless of their visual fidelity. 
Consequently, we argue that future FDeID research should report \metric\ alongside its constituent components rather than any single privacy or utility number in isolation, and we release \uface\ and the \metric\ toolkit to facilitate standardized benchmarking, reproducible evaluation, and further research in face de-identification.

\section{Conclusion}

We presented \uface\ and \metric, a paired benchmark and hierarchical metric for face de-identification.
\uface\ is a curated, demographically balanced corpus assembled from four large-scale sources. \metric\ unifies identity suppression, three-level utility preservation, and image quality into a single consistency-based score, computed without per-image attribute annotations. A comparative study over $12$ representative methods spanning adversarial, GAN-based, and diffusion-based families surfaces trade-offs invisible to existing protocols: hierarchical signals degrade asymmetrically, with macro cues mostly retained, micro cues partially degraded, and physiological rPPG silently destroyed by virtually all generative approaches; most adversarial methods, once evaluated against an ensemble of modern recognizers, exhibit limited identity suppression; and no current method is simultaneously private and utility-preserving across all three levels.

\noindent\textbf{Limitation.} 
\metric\ is consistency-based, so its validity rests on the pretrained estimators used to compute each consistency score, and any bias of these estimators is inherited by \metric. Ensembling for $\bar{P}$ and $U_3$ and the substitution study of Appendix~\ref{sec:app_estsub} bound this dependence but do not remove it: the released protocol scores L1 and L2 with one estimator per attribute, and improvements in the underlying models will translate directly into more faithful scores.

\noindent\textbf{Acknowledgements.}
This work was supported by the Research Council of Finland (former Academy of Finland) Academy Professor project EmotionAI (grants 336116, 359894), Academy project TransFace (grant number 376406), HPC project FaceCanvas (grant number 364905), the Finnish Doctoral Program Network in Artificial Intelligence (AI-DOC, decision number VN/3137/2024-OKM-6), the University of Oulu \& Research Council of Finland Profi 7 (grant 352788), and EU HORIZON-MSCA-SE-2022 project ACMod (grant 101130271). The authors also wish to acknowledge CSC – IT Center for Science, Finland, for computational resources.

{\small
\bibliographystyle{plain}

\bibliography{reference}
}

\appendix

\section*{Technical Appendix}

\section{\uface\ Benchmark: Extended Details}
\label{sec:app_benchmark}
This section completes §\ref{subsec:construction} with the operational details deferred for space: per-source statistics, the identity-deduplication threshold, the pose-conditioned routing between GFPGAN and CodeFormer, the empirical NIQE\,$<$\,5.0 cut-off, the composite ranking driving hierarchical stratified sampling, and source-dataset licenses.

\subsection{Source Dataset Statistics}
We aggregate four large-scale face recognition datasets, summarized in Tab.~\ref{tab:app_sources}.
These datasets differ substantially in scale, identity count, and image-per-identity ratio, which motivates the aggregation: no single
source provides the combination of identity diversity and image coverage required for reliable privacy evaluation.

\begin{table}[h]
\centering
\vspace{-4mm}
\caption{\textbf{Source datasets aggregated into the \uface\ pool.}}
\label{tab:app_sources}
\resizebox{0.55\columnwidth}{!}{%
\begin{tabular}{@{}lrrr@{}}
\toprule
\textbf{Dataset} & \textbf{\#Identities} & \textbf{\#Images} & \textbf{Images / ID} \\
\midrule
WebFace42M~\cite{zhu2021webface260m}       & $2$M & $42$M & $21.0$ \\
Glint360K~\cite{an2021partial}             & $360$K     & $17$M & $47.2$ \\
MS1MV2~\cite{deng2019arcface}              & $85$K      & $5.8$M  & $68.2$ \\
MegaFace~\cite{kemelmacher2016megaface}    & $672$K     & $4.7$M  & $7.1$  \\
\midrule
\textbf{Aggregated pool}                   & $\sim 3.12$M    & $\sim 69.55$M    & $22.3$ \\
\bottomrule
\end{tabular}%
}
\end{table}

\subsection{Identity-Aware Cleaning}
\label{sec:app_cleaning}

We discard samples that would compromise downstream evaluation through four criteria:
(i)~images with zero or multiple detected faces, which introduce annotation ambiguity;
(ii)~non-RGB images (\eg, grayscale, infrared), which are incompatible with color-dependent estimators, in particular rPPG;
(iii)~heavily blurred images, particularly prevalent in WebFace42M;
(iv)~cross-dataset duplicate identities, identified by computing pairwise cosine similarities between per-identity mean embeddings from
a pretrained ArcFace model~\cite{deng2019arcface} from the InsightFace library.
Identity pairs whose cosine similarity exceeds $\tau_{\text{id}} = 0.55$ are merged (when from different sources) or the lower-quality instance removed. The threshold $\tau_{\text{id}} = 0.55$ was chosen empirically: on a held-out set of manually verified same-identity and different-identity pairs, this value yielded a false-merge rate below $1\%$ while capturing over $90\%$ of cross-dataset duplicates.

\subsection{Resolution Enhancement}
\label{sec:app_enhancement}

Surviving low-resolution samples are upsampled to $448{\times}448$ or $512{\times}512$ using blind face restoration, with the upsampling
ratio capped at $4{\times}$ to preserve perceptual fidelity. We employ two complementary restoration methods:
GFPGAN~\cite{wang2021towards}, which produces high-quality results for near-frontal faces, and CodeFormer~\cite{zhou2022towards}, which is
more robust for faces with substantial pose variation where GFPGAN introduces visible artifacts such as eye-region asymmetry and
texture flattening.
Head pose is estimated from the detected $5$-point landmarks; samples with an estimated yaw magnitude above $30^\circ$ are routed to
CodeFormer, and the remainder to GFPGAN. Samples undetectable by MTCNN~\cite{zhang2016joint} after both restoration attempts are discarded.

\subsection{Perceptual Quality Filtering}
\label{sec:app_quality}

We apply a no-reference perceptual quality filter based on NIQE~\cite{mittal2012making}, discarding images with $\text{NIQE} \geq 5.0$.
The threshold was chosen empirically using FFHQ~\cite{karras2019style} as a naturalistic-face quality anchor: $95\%$ of FFHQ images fall
below $\text{NIQE} = 5.0$, providing a principled cut-off that admits naturalistic, high-fidelity faces while rejecting restoration failures
and residual low-quality samples from the source datasets.
Filtering at this threshold removes approximately $58\%$ of the pool after Stage~2, disproportionately concentrated in the heavily-blurred
WebFace42M subset.

\subsection{Hierarchical Stratified Sampling}
\label{sec:app_sampling}

After Stage~3 the pool, while large and high-quality, exhibits strong distributional skew:
males account for $66.4\%$ of detected faces, White subjects $69.5\%$, ``Happy'' expressions $51.2\%$, frontal head poses $53.3\%$, and the 45--55 age bin alone $38.2\%$. Categories such as Southeast Asian ($1.3\%$), Fear ($1.0\%$), and Strong-Right pose ($7.3\%$) are severely underrepresented (Fig.~\ref{fig:statistics}, grey bars). Evaluating FDeID methods on such a skewed distribution would conflate
method performance with demographic bias.

We correct this with hierarchical stratified sampling at the identity level.
Identities are first partitioned into $14$ strata defined by the product of two gender categories and seven ethnicity categories.
Within each stratum, identities are ranked by a composite score
\begin{equation}
    \text{score}(\text{id}) = \alpha_{\text{cnt}} \cdot \log(n_{\text{id}}) + \alpha_{\text{div}} \cdot \bigl(H_{\text{exp}}(\text{id}) + H_{\text{pose}}(\text{id})\bigr),
\end{equation}
where $n_{\text{id}}$ is the number of images belonging to the
identity, $H_{\text{exp}}$ and $H_{\text{pose}}$ are Shannon entropies
over expression and head-pose categories respectively, and
$\alpha_{\text{cnt}} = \alpha_{\text{div}} = 0.5$ by default.
The top-ranked identities fill an equal per-stratum budget of
approximately $150$ identities, and all images of each selected
identity are retained to maximize per-identity coverage.

Although not directly stratified, expression, head pose, and age achieve full coverage via diversity-aware ranking, which up-weights underrepresented categories (Fear, Disgust, Strong-Left/Right, ages 15–25 / 55–65) by $1.5$--$4\times$ over the Stage-3 pool.

\subsection{Licenses and Release}
\label{sec:app_licenses}

Source datasets are used under their respective licenses:
WebFace42M (research-only, non-commercial),
Glint360K (research-only),
MS1MV2 (research-only, derived from MS-Celeb-1M),
MegaFace (research-only, subject to the MegaFace Terms of Use).
Each of these licenses permits research use and derivative processing of the data, and none forbids the construction or distribution of derived identifiers such as identity keys and file lists.
We release \uface\ as a set of identity-indexed processing and sampling scripts together with the list of selected identity keys and image
filenames from each source, rather than as a redistributed image archive, so that downstream users obtain the underlying images
directly from the original providers under the original licenses and the release exposes nothing beyond the public sources. The release scripts, evaluation toolkit, and computed \metric\ scores for all baseline methods are made available for research use.

\paragraph{Consent and provenance.}
All sources were collected by their original authors from public web imagery of (mostly public) figures without individual consent, and MS1MV2 in particular derives from MS-Celeb-1M, whose collection has drawn community scrutiny. We acknowledge this explicitly. 

\paragraph{Safeguards.}
The identity keys are opaque per-source indices and are useful only to someone who already holds the licensed source data. The release is gated behind the same research-only terms as the sources, the accompanying documentation states the intended use (evaluation of de-identification, not training of recognizers, not re-identification of individuals).

\paragraph{Broader impact.}
The intended positive impact is a common, annotation-free yardstick that lets privacy-preserving face analysis be compared and improved reproducibly, and that exposes silent failures (\eg, destroyed physiological signals) before deployment in telehealth or surveillance contexts. Two negative uses are foreseeable. First, a benchmark that scores identity suppression against modern recognizers can be used to tune methods whose purpose is to evade lawful face recognition; we consider this risk modest because the recognizers we ensemble are public and the adversarial methods we evaluate already exist, and because the same measurements are what defenders need. Second, any face benchmark concentrates biometric data; the release design above (no pixels, no embeddings, gated research-only terms, removal on request) is our mitigation. \metric\ is a machine-recognizer score and should not be read as a guarantee of anonymity against human observers or future recognizers.

\section{\metric\ Metric: Extended Details}
\label{sec:app_metric}

§\ref{sec:fdeidscore} named \metric's consistency functions and stated the resulting properties without giving their explicit forms. This section supplies them: the closed form of every L1, L2, and L3 consistency function; the global normalization constants and their rationale; and an aggregation ablation establishing the weighted harmonic mean as the only choice consistent with the vetoing property.

\subsection{Consistency Function Definitions}
\label{sec:app_metric_agreements}

Given an original face $\mathbf{x}$ (or video clip $\mathbf{X}$) and
its de-identified counterpart $\hat{\mathbf{x}}$ (or $\hat{\mathbf{X}}$),
every consistency function is defined as follows.

\paragraph{Age estimation (L1).}
The estimator $f_{\text{age}}(\cdot)$ outputs a scalar age
$a \in [0, 100]$; we use truncated absolute error:
\begin{equation}
    s_{\text{age}}(\mathbf{x}, \hat{\mathbf{x}}) = \max\!\left(0,\; 1 - \frac{|f_{\text{age}}(\mathbf{x}) - f_{\text{age}}(\hat{\mathbf{x}})|}{\Delta_{\text{age}}}\right),
    \quad \Delta_{\text{age}} = 100.
\end{equation}

\paragraph{Ethnicity and macro-expression recognition (L1).}
Both tasks produce softmax outputs $\mathbf{p} \in \Delta^{K-1}$ over $K$ classes ($K=7$ for FairFace ethnicity, $K=7$ for POSTER macro-expression on RAF-DB).
Rather than using only the top-1 label, we use total variation distance between the full posterior distributions, which captures both the label
match and the confidence profile:
\begin{equation}
  s_{\text{cls}}(\mathbf{x}, \hat{\mathbf{x}}) = 1 - \tfrac{1}{2}\,\bigl\|f_{\text{cls}}(\mathbf{x}) -
f_{\text{cls}}(\hat{\mathbf{x}})\bigr\|_1 \;\in\; [0, 1],
\end{equation}
for $\text{cls} \in \{\text{ethnicity}, \text{macro-exp}\}$.
Two methods producing the same top-1 class but different posterior distributions are distinguished correctly under this definition.

\paragraph{Gender recognition (L1).}
The MiVOLO age and gender estimator emits a top-1 binary label
$\hat{y} \in \{\text{male},\text{female}\}$ and an associated confidence
$c \in [0, 1]$ rather than a softmax head. We construct an equivalent
binary posterior $\mathbf{p}_{\text{gender}} \in \Delta^{1}$ by placing
mass $c$ on the predicted class and $1{-}c$ on the other, and then apply
the same total-variation consistency:
\begin{equation}
  \mathbf{p}_{\text{gender}} =
  \begin{cases}
    [c,\; 1-c] & \hat{y} = \text{male}, \\
    [1-c,\; c] & \hat{y} = \text{female},
  \end{cases}
  \qquad
  s_{\text{gender}}(\mathbf{x}, \hat{\mathbf{x}}) = 1 - \tfrac{1}{2}\,\bigl\|\mathbf{p}_{\text{gender}}(\mathbf{x})
- \mathbf{p}_{\text{gender}}(\hat{\mathbf{x}})\bigr\|_1.
\end{equation}
This reduces to a indicator weighted by the joint confidence: $s_{\text{gender}} = 1$ when both predictions agree with full confidence and decreases linearly as either confidence drops or the labels diverge.

\paragraph{Landmark detection (L1).}
The estimator $f_{\text{lmk}}(\cdot)$ outputs $K_{\text{lmk}}$ 2D
coordinates $\mathbf{L} \in \mathbb{R}^{K_{\text{lmk}} \times 2}$.
We quantify consistency via the normalized mean error (NME), using the
inter-ocular distance $d_{\text{io}}$ measured on $\mathbf{x}$ (not
$\hat{\mathbf{x}}$) as normalizer:
\begin{equation}
    \text{NME}(\mathbf{L}, \hat{\mathbf{L}}) = \frac{1}{K_{\text{lmk}}\, d_{\text{io}}} \sum_{i=1}^{K_{\text{lmk}}} \|\mathbf{L}_i - \hat{\mathbf{L}}_i\|_2,
    \quad
    s_{\text{lmk}}(\mathbf{x}, \hat{\mathbf{x}}) = \max\!\left(0,\; 1 - \frac{\text{NME}}{\tau_{\text{NME}}}\right),
\end{equation}
with cap $\tau_{\text{NME}} = 0.10$, beyond which face landmarks are
considered unreliable in standard benchmarks.
Anchoring $d_{\text{io}}$ on the original image ensures that any scale
changes introduced by de-identification (\eg, face warping, region
replacement) are properly penalized.

\paragraph{L1 aggregation.}
The L1 utility score is the unweighted mean of the five consistency scores:
\begin{equation}
    U_1(\mathbf{x}, \hat{\mathbf{x}}) = \tfrac{1}{5}\bigl(s_{\text{age}} + s_{\text{gender}} + s_{\text{eth}} + s_{\text{macro-exp}} + s_{\text{lmk}}\bigr).
\end{equation}

\paragraph{Gaze estimation (L2).}
The estimator $f_{\text{gaze}}$ outputs a 3D unit gaze direction
$\mathbf{g} \in \mathbb{S}^2$.
We use the standard angular error, normalized by a
$\theta_{\max} = 90^\circ$ cap:
\begin{equation}
    \theta(\mathbf{g}, \hat{\mathbf{g}}) = \arccos\bigl(\mathbf{g} \cdot \hat{\mathbf{g}}\bigr),
    \quad
    s_{\text{gaze}}(\mathbf{x}, \hat{\mathbf{x}}) = \max\!\left(0,\; 1 - \frac{\theta(\mathbf{g}, \hat{\mathbf{g}})}{\theta_{\max}}\right).
\end{equation}

\paragraph{Micro-expression recognition (L2).}
This task is video-based.
Let $\mathbf{X}$ and $\hat{\mathbf{X}}$ denote the original and
de-identified video clips, and let $f_{\text{me}}$ output a softmax
distribution over ME classes.
The class space differs from macro-expression: SAMER~\cite{ruan2022mimicking} outputs a posterior over the five CASME~II categories (happiness, disgust, repression, surprise, others), and we apply it to DFME clips without retraining. The consistency function is shared with the L1 categorical tasks:
\begin{equation}
    s_{\text{me}}(\mathbf{X}, \hat{\mathbf{X}}) = 1 - \tfrac{1}{2}\,\bigl\|f_{\text{me}}(\mathbf{X}) - f_{\text{me}}(\hat{\mathbf{X}})\bigr\|_1.
\end{equation}

\paragraph{L2 aggregation.}
For image-only inputs where $s_{\text{me}}$ is undefined, $U_2$
reduces to gaze alone:
\begin{equation}
    U_2 = \begin{cases}
        \tfrac{1}{2}\bigl(s_{\text{gaze}} + s_{\text{me}}\bigr) & \text{if video}, \\
        s_{\text{gaze}} & \text{if image only}.
    \end{cases}
\end{equation}
Because $U_2$ integrates sub-scores across modalities, we report
$s_{\text{gaze}}$ and $s_{\text{me}}$ separately in experimental
tables and only aggregate into $U_2$ at the final composite step, so
that cross-modality comparisons remain interpretable.

\paragraph{rPPG signals (L3).}
L3 is intrinsically video-level. Given an rPPG estimator $h(\cdot)$, we obtain a BVP waveform $\mathbf{b} \in \mathbb{R}^{T}$ and a derived heart rate $r$ in beats per minute (bpm).
Because destroyed rPPG signals typically manifest as \emph{uncorrelated}
output rather than anti-correlated, we use a positive-correlation
rescaling for BVP, assigning zero consistency to uncorrelated or
anti-correlated pairs; HR uses the tightened clinical-relevance cap
$\tau_{\text{HR}} = 20$\,bpm:
\begin{equation}
    s_{\text{BVP}}(\mathbf{X}, \hat{\mathbf{X}}) = \max\!\bigl(0,\; \rho(\mathbf{b}, \hat{\mathbf{b}})\bigr),
    \qquad
    s_{\text{HR}}(\mathbf{X}, \hat{\mathbf{X}}) = \max\!\left(0,\; 1 - \frac{|r - \hat{r}|}{\tau_{\text{HR}}}\right).
\end{equation}
By convention, $\rho = 0$ whenever either signal has zero variance.
To reduce estimator-specific bias, we average over an ensemble of
$K_3$ rPPG estimators:
\begin{equation}
    U_3 = \frac{1}{K_3} \sum_{k=1}^{K_3} \tfrac{1}{2}\bigl(s_{\text{BVP}}^{(k)} + s_{\text{HR}}^{(k)}\bigr).
\end{equation}

\paragraph{Privacy score.}
Identity suppression is measured as an embedding-space dissimilarity
under a face recognition model $\mathcal{R}$.
Because cosine similarity takes values in $[-1, 1]$, we rescale to
$[0, 1]$:
\begin{equation}
    P(\mathbf{x}, \hat{\mathbf{x}}) = \tfrac{1}{2}\bigl(1 - \cos(\mathcal{R}(\mathbf{x}), \mathcal{R}(\hat{\mathbf{x}}))\bigr).
\end{equation}
To reduce dependence on any single backbone, a known vulnerability
when evaluating adversarial methods optimized against specific
recognizers, we average across an ensemble of $M$ recognition models:
$\bar{P} = \frac{1}{M} \sum_{m=1}^{M} P_m$.

\paragraph{Quality score.}
We combine a reference-based perceptual similarity and a no-reference
naturalness score, with both terms bounded to $[0, 1]$:
\begin{equation}
    Q(\mathbf{x}, \hat{\mathbf{x}}) = \alpha \cdot \bigl(1 - \min(\text{LPIPS}(\mathbf{x}, \hat{\mathbf{x}}),\, 1)\bigr) + (1 - \alpha) \cdot \text{NIQE}^{*}(\hat{\mathbf{x}}),
\end{equation}
where $\text{LPIPS}$~\cite{zhang2018unreasonable} captures perceptual
fidelity and
$\text{NIQE}^{*}(\hat{\mathbf{x}}) = \max(0, \min(1, (\tau_{\text{N}} - \text{NIQE}(\hat{\mathbf{x}}))/\tau_{\text{N}}))$
is the normalized inverse NIQE score with $\tau_{\text{N}} = 5$,
matching our curation threshold.
LPIPS values exceed $1$ for heavily corrupted inputs;
clipping via $\min(\cdot, 1)$ is inconsequential in practice because
such cases already correspond to visually destroyed faces.
We set $\alpha = 0.5$ by default.

\subsection{Normalization-Constant Choices}
\label{sec:app_metric_caps}

Tab.~\ref{tab:app_caps} summarizes the global normalization constants
used throughout \metric\ and the rationale for each.
These constants are specified a priori and shared across all methods,
so they do not introduce method-specific bias.

\begin{table}[h]
\centering
\vspace{-4mm}
\caption{\textbf{Normalization constants in \metric.}}
\label{tab:app_caps}
\resizebox{0.75\columnwidth}{!}{%
\begin{tabular}{@{}llp{8cm}@{}}
\toprule
\textbf{Constant} & \textbf{Value} & \textbf{Rationale} \\
\midrule
$\Delta_{\text{age}}$ & $100$ years & Full biological age range. \\
$\tau_{\text{NME}}$ & $0.10$ & Threshold beyond which face landmarks are considered unreliable in standard benchmarks. \\
$\theta_{\max}$ & $90^\circ$ & Maximum possible angular error between two 3D unit vectors in the forward hemisphere. \\
$\tau_{\text{HR}}$ & $20$\,bpm & Clinical relevance threshold; errors above this render HR estimates unusable for downstream tasks. \\
$\tau_{\text{N}}$ & $5$ & Matches the NIQE curation threshold used to filter \uface. \\
$\alpha$ & $0.5$ & Equal weighting of reference-based and no-reference quality components. \\
$\epsilon$ & $10^{-6}$ & Prevents division by zero in the harmonic mean. \\
\bottomrule
\end{tabular}%
}
\end{table}

\subsection{Aggregation-Choice Ablation}
\label{sec:app_metric_aggregation}

We compare three aggregation schemes on the same five component scores $\{\bar{P}, Q, U_1, U_2, U_3\}$ under uniform weights, to justify the weighted harmonic mean as the choice for \metric.

\paragraph{Aggregation formulas.}
Let $\mathbf{c} = (\bar{P},\, Q,\, U_1,\, U_2,\, U_3) \in [0,1]^5$ be the per-method component vector, and let $n = 5$ denote its dimensionality.
The three schemes evaluated under uniform weights $w_i = 1/n$ are
\begin{equation}
\begin{aligned}
\mathcal{A}(\mathbf{c}) &= \tfrac{1}{n}\sum_{i=1}^{n} c_i
  && \text{(arithmetic mean)}, \\
\mathcal{G}(\mathbf{c}) &= \Bigl(\textstyle\prod_{i=1}^{n} c_i\Bigr)^{1/n}
  && \text{(geometric mean)}, \\
\mathcal{H}(\mathbf{c}) &= n \,\Big/\, \textstyle\sum_{i=1}^{n} \dfrac{1}{c_i + \epsilon}
  && \text{(weighted harmonic mean, our \metric)},
\end{aligned}
\label{eq:app_aggregation}
\end{equation}
with $\epsilon = 10^{-6}$. 
$\mathcal{H}$ in Eq.~\eqref{eq:app_aggregation} is the uniform-weight specialization of the weighted form in Eq.~\eqref{eq:fdeid}: substituting $w_P = w_Q = w_1 = w_2 = w_3 = 1/5$ into the numerator $\sum w$ gives $1$ and the denominator becomes $\frac{1}{n}\sum_i \frac{1}{c_i+\epsilon}$, which after multiplying numerator and denominator by $n$ yields $\mathcal{H}$.
The known inequality $\mathcal{H} \le \mathcal{G} \le \mathcal{A}$ (with equality only when all $c_i$ are identical) means the three schemes always agree on \emph{perfectly balanced} methods but diverge sharply when one $c_i$ is near zero.

\paragraph{Worked example (Adv-Makeup).}
We illustrate on Adv-Makeup, whose component vector is
$\mathbf{c} = (0.012,\, 0.524,\, 0.960,\, 0.941,\, 0.931)$:
\begin{align*}
\mathcal{A} &= (0.012 + 0.524 + 0.960 + 0.941 + 0.931)/5  \approx 0.673, \\
\mathcal{G} &= (0.012 \times 0.524 \times 0.960 \times 0.941 \times 0.931)^{1/5}  \approx 0.350, \\
\mathcal{H} &= 5 \,\big/\, \bigl(\tfrac{1}{0.012{+}\epsilon} + \tfrac{1}{0.524} + \tfrac{1}{0.960} + \tfrac{1}{0.941} + \tfrac{1}{0.931}\bigr) \approx 0.056.
\end{align*}
The single near-zero component $\bar{P}=0.012$ is invisible to $\mathcal{A}$, mildly visible to $\mathcal{G}$, and dominates $\mathcal{H}$ via the $1/0.012\!\approx\!83$ term that swamps the other four reciprocals (each $\le 1/0.524 \approx 1.91$).

\begin{table}[h]
\centering
\caption{\textbf{Aggregation-scheme comparison on the Balanced profile (all 12 methods).}
Each cell is computed from the same five component scores $\{\bar{P}, Q, U_1, U_2, U_3\}$ in Tab.~\ref{tab:hifd_main} via Eq.~\eqref{eq:app_aggregation} with uniform weights.
Ranks (gray) are over the full set of twelve methods.
}
\label{tab:app_aggregation}
\resizebox{0.92\textwidth}{!}{%
\begin{tabular}{@{}lccccc ccc@{}}
\toprule
\multirow{2}{*}{\textbf{Method}} & \multicolumn{5}{c}{\textbf{Component scores}} & \multicolumn{3}{c}{\textbf{Aggregation scheme (value, rank / 12)}} \\
\cmidrule(lr){2-6} \cmidrule(lr){7-9}
 & $\bar{P}$ & $Q$ & $U_1$ & $U_2$ & $U_3$ & $\mathcal{A}$ \textbf{(Arith.)} & $\mathcal{G}$ \textbf{(Geom.)} & $\mathcal{H}$ \textbf{(Harm., Ours)} \\
\midrule
\multicolumn{9}{l}{\textit{(a) Adversarial perturbation}} \\
MI-FGSM       & $0.134$ & $0.360$ & $0.928$ & $0.868$ & $0.474$ & $0.553$ \textcolor{gray}{(8)}  & $0.450$ \textcolor{gray}{(7)}  & $0.343$ \textcolor{gray}{(7)}  \\
PGD           & $0.113$ & $0.362$ & $0.938$ & $0.878$ & $0.566$ & $0.572$ \textcolor{gray}{(5)}  & $0.453$ \textcolor{gray}{(6)}  & $0.321$ \textcolor{gray}{(9)}  \\
TI-DIM        & $0.500$ & $0.342$ & $0.757$ & $0.821$ & $0.424$ & $0.569$ \textcolor{gray}{(6)}  & $0.538$ \textcolor{gray}{(3)}  & $0.509$ \textcolor{gray}{(2)}  \\
TIP-IM        & $0.496$ & $0.446$ & $0.806$ & $0.878$ & $0.624$ & $0.650$ \textcolor{gray}{(3)}  & $\mathbf{0.628}$ \textcolor{gray}{(1)} & $\mathbf{0.607}$ \textcolor{gray}{(1)} \\
Chameleon     & $0.239$ & $0.474$ & $0.850$ & $0.889$ & $0.799$ & $0.650$ \textcolor{gray}{(2)}  & $0.585$ \textcolor{gray}{(2)}  & $0.508$ \textcolor{gray}{(3)}  \\
\midrule
\multicolumn{9}{l}{\textit{(b) GAN-based generation}} \\
Adv-Makeup    & $0.012$ & $0.524$ & $0.960$ & $0.941$ & $0.931$ & $\mathbf{0.673}$ \textcolor{gray}{(1)} & $0.350$ \textcolor{gray}{(11)} & $0.056$ \textcolor{gray}{(12)} \\
AMT-GAN       & $0.156$ & $0.481$ & $0.861$ & $0.745$ & $0.148$ & $0.478$ \textcolor{gray}{(11)} & $0.372$ \textcolor{gray}{(10)} & $0.282$ \textcolor{gray}{(10)} \\
CIAGAN        & $0.450$ & $0.276$ & $0.468$ & $0.660$ & $0.247$ & $0.420$ \textcolor{gray}{(12)} & $0.394$ \textcolor{gray}{(9)}  & $0.369$ \textcolor{gray}{(6)}  \\
DeID-rPPG     & $0.013$ & $0.407$ & $0.932$ & $0.889$ & $0.907$ & $0.630$ \textcolor{gray}{(4)}  & $0.332$ \textcolor{gray}{(12)} & $0.061$ \textcolor{gray}{(11)} \\
G$^{2}$Face   & $0.210$ & $0.443$ & $0.809$ & $0.848$ & $0.326$ & $0.527$ \textcolor{gray}{(9)}  & $0.461$ \textcolor{gray}{(5)}  & $0.400$ \textcolor{gray}{(4)}  \\
\midrule
\multicolumn{9}{l}{\textit{(c) Diffusion-based generation}} \\
WeakenDiff    & $0.222$ & $0.371$ & $0.834$ & $0.792$ & $0.185$ & $0.481$ \textcolor{gray}{(10)} & $0.398$ \textcolor{gray}{(8)}  & $0.332$ \textcolor{gray}{(8)}  \\
DiffAM        & $0.159$ & $0.471$ & $0.793$ & $0.811$ & $0.557$ & $0.558$ \textcolor{gray}{(7)}  & $0.485$ \textcolor{gray}{(4)}  & $0.394$ \textcolor{gray}{(5)}  \\
\bottomrule
\end{tabular}%
}
\end{table}

\paragraph{Discussion.}
Three failure modes of the non-harmonic schemes emerge from Tab.~\ref{tab:app_aggregation}.
\textit{(i)~Arithmetic mean is uninformative for unbalanced profiles.}
Adv-Makeup ($\bar{P}\!=\!0.012$) achieves the \emph{highest} $\mathcal{A} = 0.673$ across all twelve methods, and DeID-rPPG ($\bar{P}\!=\!0.013$) ranks fourth at $\mathcal{A} = 0.630$, both near-zero-privacy methods are placed in the top four despite offering essentially no identity suppression.
The valid leader TIP-IM is demoted to rank $3$ ($\mathcal{A} = 0.650$), trailing Adv-Makeup by $0.023$.
\textit{(ii)~Geometric mean ranks the bottom correctly but scores it too leniently.} Adv-Makeup and DeID-rPPG, both with near-zero privacy ($\bar{P}\!\leq\!0.013$), correctly fall to ranks $11$ and $12$ under $\mathcal{G}$, yet their scores ($\mathcal{G}\!=\!0.350$ and $0.332$) land within $0.05$ of mid-pack AMT-GAN ($\mathcal{G}\!=\!0.372$), whose privacy is an order of magnitude higher ($\bar{P}\!=\!0.156$). Under $\mathcal{H}$, the same two methods score $4$ to $5$ times lower than AMT-GAN ($\mathcal{H}\!=\!0.282$), and the leader-to-loser gap widens from $\mathcal{G}$'s $0.296$ to $0.551$. Vetoing requires that a near-zero component drive the composite toward zero, not merely to the bottom of a narrow range, so $\mathcal{G}$ satisfies vetoing in rank but not in score.
\textit{(iii)~Harmonic mean correctly resolves both pathologies.}
Adv-Makeup and DeID-rPPG drop to ranks $12$ and $11$ at $\mathcal{H} \in \{0.056, 0.061\}$, an order of magnitude below the leader TIP-IM ($\mathcal{H} = 0.607$); the gap is now $0.551$.
The rank order between TIP-IM (rank 1) and Adv-Makeup (rank 12) under $\mathcal{H}$ is exactly inverted relative to $\mathcal{A}$ (where Adv-Makeup is rank 1 and TIP-IM rank 3): the harmonic mean does not merely widen an existing gap, it reverses an incorrect ranking.

\paragraph{Why this matters.}
Property (iii) \emph{vetoing} from §\ref{sec:fdeidscore} requires that any near-zero component drives the composite toward zero (exactly what is observed for $\mathcal{H}$ on Adv-Makeup and DeID-rPPG).
This property is what makes \metric\ a \emph{valid} privacy-utility trade-off score: an FDeID method that fails on identity suppression should not be ranked highly regardless of its utility profile, because such a method does not de-identify in the first place.
Neither $\mathcal{A}$ (which inverts the correct ranking by placing Adv-Makeup at rank $1$) nor $\mathcal{G}$ (which preserves the ranking but compresses near-zero-privacy methods into the same neighborhood as methods with real identity suppression) satisfies this requirement, justifying the weighted harmonic mean as the only sound choice for \metric.

\section{Validity, Uncertainty, and Robustness of \metric}
\label{sec:app_robustness}
This section collects the experiments that establish how far \metric\ can be trusted: the validation of the consistency paradigm behind §\ref{sec:validity} against ground truth, the operational calibration of $\bar{P}$, statistical uncertainty, estimator substitution, weight sensitivity, and the quality axis.

\subsection{Validation of the Consistency Paradigm}
\label{sec:app_validation}
The consistency formulation replaces per-image labels with estimator consistency. This section tests, on the two attribute families for which ground truth is obtainable, whether consistency tracks true utility preservation; §\ref{sec:validity} summarizes the results.

\subsubsection{Macro-Expression: RAF-DB Test Split}
We de-identified the entire RAF-DB test split ($3{,}068$ images with ground-truth seven-class expression labels; POSTER's training distribution, which is what makes labels available) with all twelve methods, and for every (method, image) pair recorded both the consistency score $s_{\text{macro-exp}}(\mathbf{x},\hat{\mathbf{x}})$ and whether the same estimator still predicts the correct label on $\hat{\mathbf{x}}$. POSTER reaches $77.8\%$ accuracy on the originals. Tab.~\ref{tab:app_rafdb} reports per-method results; pooled over $28{,}614$ pairs, the consistency score predicts label survival with AUC $0.981$, and among the $20{,}096$ pairs with consistency above $0.8$ only $1.06\%$ have a flipped true label. Across methods, mean consistency correlates with ground-truth accuracy retention at Spearman $0.944$ and Pearson $0.992$; the leading group (Adv-Makeup, DeID-rPPG, AMT-GAN) and the worst method (CIAGAN) coincide under both. The out-of-distribution complement is the POSTER/AffectNet substitution of Appendix~\ref{sec:app_estsub}, which agrees with the RAF-DB-trained estimator at Spearman $0.979$.

\begin{table}[h]
\centering
\caption{\textbf{Does consistency track true expression preservation? (RAF-DB test split, $n{=}3{,}068$).} Consistency is the mean $s_{\text{macro-exp}}$; accuracy is POSTER's top-1 accuracy on the de-identified images against ground truth (originals: $0.778$), with a 95\% image-level bootstrap CI; retention is accuracy on $\hat{\mathbf{x}}$ over accuracy on $\mathbf{x}$; AUC is the per-image ability of the consistency score to predict that the true label survives; the last column is the fraction of pairs with consistency $>0.8$ whose true label nevertheless flipped.}
\label{tab:app_rafdb}
\resizebox{1.0\textwidth}{!}{%
\begin{tabular}{@{}l c l c c c@{}}
\toprule
\textbf{Method} & \textbf{Consistency} & \textbf{Accuracy on $\hat{\mathbf{x}}$} & \textbf{Retention} & \textbf{AUC} & \textbf{Hidden flips} \\
\midrule
\multicolumn{6}{l}{\textit{(a) Adversarial perturbation}} \\
MI-FGSM       & 0.813 & 0.646 {\scriptsize[0.629,\,0.662]} & 0.830 & 0.976 & 1.7\% \\
PGD           & 0.860 & 0.694 {\scriptsize[0.678,\,0.710]} & 0.892 & 0.971 & 1.6\% \\
TI-DIM        & 0.777 & 0.625 {\scriptsize[0.607,\,0.642]} & 0.804 & 0.979 & 0.7\% \\
TIP-IM        & 0.779 & 0.613 {\scriptsize[0.597,\,0.630]} & 0.788 & 0.983 & 0.8\% \\
Chameleon     & 0.792 & 0.621 {\scriptsize[0.604,\,0.638]} & 0.798 & 0.974 & 1.1\% \\
\midrule
\multicolumn{6}{l}{\textit{(b) GAN-based generation}} \\
Adv-Makeup    & 0.970 & 0.777 {\scriptsize[0.762,\,0.792]} & 0.999 & 0.968 & 1.3\% \\
AMT-GAN       & 0.867 & 0.710 {\scriptsize[0.694,\,0.725]} & 0.912 & 0.981 & 0.5\% \\
CIAGAN        & 0.509 & 0.370 {\scriptsize[0.353,\,0.387]} & 0.476 & 0.980 & 0.2\% \\
DeID-rPPG     & 0.948 & 0.781 {\scriptsize[0.766,\,0.795]} & 1.003 & 0.974 & 1.4\% \\
G$^{2}$Face   & 0.815 & 0.671 {\scriptsize[0.654,\,0.687]} & 0.863 & 0.975 & 1.0\% \\
\midrule
\multicolumn{6}{l}{\textit{(c) Diffusion-based generation}} \\
WeakenDiff    & 0.795 & 0.663 {\scriptsize[0.646,\,0.679]} & 0.852 & 0.980 & 0.7\% \\
DiffAM        & 0.797 & 0.647 {\scriptsize[0.631,\,0.664]} & 0.832 & 0.976 & 0.9\% \\
\midrule
\textit{Pooled / across methods} & \multicolumn{5}{l}{AUC $0.981$ over $28{,}614$ pairs; hidden flips $1.06\%$ of $20{,}096$; Spearman(consistency, retention) $= 0.944$, Pearson $= 0.992$} \\
\bottomrule
\end{tabular}%
}
\end{table}

\paragraph{When consistency can fail.}
For categorical attributes the consistency score is $1 - \mathrm{TV}(\mathbf{p}, \hat{\mathbf{p}})$, and total variation upper-bounds the change in probability mass on any class, including the true one, so a large drop in \emph{soft} ground-truth accuracy cannot coexist with high consistency. The \emph{hard-label} case is not bounded in the worst case: $\mathbf{p} = (0.51, 0.49)$ and $\hat{\mathbf{p}} = (0.49, 0.51)$ have $\mathrm{TV} = 0.02$ and a certain argmax flip. This near-tie case is why we measured hidden flips rather than assumed them away. The second failure mode is an estimator that is degenerate on the de-identified domain and therefore agrees with itself trivially (the ``poorly trained estimator outputs neutral for both'' scenario). Independently trained models do not share a degenerate failure mode, so Appendix~\ref{sec:app_estsub} tests this case and finds sub-score correlations of $0.94$ to $0.98$ across estimators.

\subsubsection{rPPG: Contact-Sensor Ground Truth}
\label{sec:app_rppg_truth}
MMPD and PURE ship contact-sensor pulse waveforms, so the true heart rate of every clip is known. For each method we compute the mean absolute HR error of the rPPG ensemble on the de-identified clips minus that on the originals ($\Delta$MAE, in bpm); a method that preserves the pulse leaves $\Delta$MAE near zero. Ranking methods by $-\Delta$MAE agrees with the consistency-based $U_3$ at Spearman $0.944$ on MMPD ($0.993$ on PURE) with the same top three (Adv-Makeup, DeID-rPPG, Chameleon) and bottom three (AMT-GAN, CIAGAN, WeakenDiff) on both datasets (per-method values in Tab.~\ref{tab:app_mmpd}). Among clips whose consistency score exceeds $0.8$, none has a de-identification-induced HR error above $10$\,bpm on either dataset ($21{,}951$ chunks on MMPD, $9{,}839$ on PURE), so a high $U_3$ never hides a destroyed pulse. In absolute terms the estimators, used without retraining, are weak on MMPD ($34.3$\,bpm MAE on the originals against $21.9$\,bpm on PURE); the consistency score measures the \emph{change} induced by de-identification, which is what the contact sensor validates, not absolute accuracy.

\subsection{Operational Privacy at Fixed False-Accept Rates}
\label{sec:app_far}
$\bar{P}$ is a rescaled embedding dissimilarity. To give it an operational reading we fit verification thresholds for each recognizer and for the ensemble score on $3$M different-identity impostor pairs drawn from \uface\ originals, at false-accept rates of $10^{-2}$, $10^{-3}$, and $10^{-4}$ (ensemble cosine thresholds $0.172$, $0.247$, and $0.316$; per-recognizer thresholds $0.165$ to $0.355$). De-identification success is the fraction of \uface\ images whose de-identified version is \emph{not} verified as its own original at that threshold. Tab.~\ref{tab:app_far} reports the ensemble success rates at all three FARs; per-recognizer rates follow the same pattern, with AdaFace the hardest recognizer to escape (\eg, TIP-IM $0.759$ / $0.868$ / $0.933$ on AdaFace against $0.930$ / $0.970$ / $0.987$ on ArcFace at the three FARs). In practitioner units, $\bar{P}\!\approx\!0.5$ maps to $93$ to $97\%$ de-identification at FAR $\le 10^{-3}$, the middle band ($\bar{P}$ from $0.11$ to $0.24$) to $27\%$ down to under $1\%$, and $\bar{P}\!\approx\!0.01$ to $0\%$; the residual (MI-FGSM and PGD reach $8\%$ at a $\bar{P}$ near AMT-GAN's) is what a continuous margin meeting a thresholded decision produces. Spearman correlation between $\bar{P}$ and success is $0.783$, $0.769$, and $0.804$ at the three thresholds; the same three methods (TI-DIM, TIP-IM, CIAGAN) lead under $\bar{P}$ and under success, their internal order differing at $10^{-3}$ and $10^{-4}$. The threat model is a machine recognizer: Appendix~\ref{sec:app_human} shows that human same-person judgments reproduce this ranking within a paradigm but not across paradigms, because adversarial perturbations are engineered against recognizers and leave the face plainly recognizable to a person.

\begin{table}[h]
\centering
\caption{\textbf{De-identification success at fixed false-accept rates} (recognizer ensemble; fraction of \uface\ faces no longer verified as their original), ordered by $\bar{P}$.}
\label{tab:app_far}
\resizebox{0.6\textwidth}{!}{%
\begin{tabular}{@{}l c ccc@{}}
\toprule
\textbf{Method} & $\bar{P}$ & \textbf{FAR} $10^{-2}$ & \textbf{FAR} $10^{-3}$ & \textbf{FAR} $10^{-4}$ \\
\midrule
TI-DIM        & 0.500 & 0.859 & 0.935 & 0.971 \\
TIP-IM        & 0.496 & 0.856 & 0.930 & 0.967 \\
CIAGAN        & 0.450 & 0.839 & 0.959 & 0.987 \\
Chameleon     & 0.239 & 0.188 & 0.270 & 0.351 \\
WeakenDiff    & 0.222 & 0.000 & 0.001 & 0.008 \\
G$^{2}$Face   & 0.210 & 0.002 & 0.010 & 0.033 \\
DiffAM        & 0.159 & 0.001 & 0.004 & 0.008 \\
AMT-GAN       & 0.156 & 0.000 & 0.002 & 0.007 \\
MI-FGSM       & 0.134 & 0.077 & 0.080 & 0.084 \\
PGD           & 0.113 & 0.078 & 0.079 & 0.081 \\
DeID-rPPG     & 0.013 & 0.000 & 0.000 & 0.000 \\
Adv-Makeup    & 0.012 & 0.000 & 0.000 & 0.000 \\
\midrule
Spearman with $\bar{P}$ & -- & 0.783 & 0.769 & 0.804 \\
\bottomrule
\end{tabular}%
}
\end{table}

\subsection{Statistical Uncertainty}
\label{sec:app_uncertainty}

\paragraph{Protocol.}
Image-level sub-scores on \uface\ are resampled at the \emph{identity} level ($2{,}062$ identity blocks over the $2{,}069$ curated identities, recovered as contiguous key runs), micro-expression at the \emph{subject} level on DFME ($42$ subjects), and rPPG at the subject level on PURE ($10$ subjects) and MMPD ($33$ subjects). Each of $B = 2{,}000$ resamples draws blocks with replacement, recomputes every component mean, and re-aggregates the composite; we report percentile 95\% intervals. Intervals therefore capture sampling variability over identities and subjects, not estimator or seed variability.

\begin{table}[h]
\centering
\caption{\textbf{Component scores with 95\% bootstrap intervals (Balanced profile).} Intervals are percentile intervals of the mean under identity-level resampling on \uface\ ($\bar{P}$, $Q$, gaze) and subject-level resampling on DFME (ME, $U_2$) and PURE ($U_3$); \metric\ combines all three. Point estimates for gaze, ME, and $U_2$ are the identity- and subject-weighted means at the center of the resampling distribution (every identity or subject counts once), which differ from the image- and clip-weighted means reported in Tab.~\ref{tab:hifd_main} by at most $0.025$. The two weightings agree on every ranking statement in the paper.}
\label{tab:app_boot}
\resizebox{\textwidth}{!}{%
\begin{tabular}{@{}l lllllll@{}}
\toprule
\textbf{Method} & $\bar{P}$ & $Q$ & Gaze & ME & $U_2$ & $U_3$ (PURE) & \metric \\
\midrule
\multicolumn{8}{l}{\textit{(a) Adversarial perturbation}} \\
MI-FGSM       & 0.134 {\scriptsize[0.133,\,0.138]} & 0.360 {\scriptsize[0.359,\,0.360]} & 0.937 {\scriptsize[0.936,\,0.938]} & 0.800 {\scriptsize[0.782,\,0.817]} & 0.868 {\scriptsize[0.859,\,0.877]} & 0.474 {\scriptsize[0.392,\,0.505]} & 0.343 {\scriptsize[0.334,\,0.348]} \\
PGD           & 0.113 {\scriptsize[0.112,\,0.117]} & 0.362 {\scriptsize[0.362,\,0.363]} & 0.941 {\scriptsize[0.939,\,0.942]} & 0.818 {\scriptsize[0.797,\,0.836]} & 0.879 {\scriptsize[0.869,\,0.889]} & 0.566 {\scriptsize[0.485,\,0.602]} & 0.321 {\scriptsize[0.316,\,0.328]} \\
TI-DIM        & 0.500 {\scriptsize[0.500,\,0.506]} & 0.342 {\scriptsize[0.341,\,0.342]} & 0.923 {\scriptsize[0.922,\,0.924]} & 0.715 {\scriptsize[0.688,\,0.740]} & 0.819 {\scriptsize[0.806,\,0.832]} & 0.424 {\scriptsize[0.356,\,0.444]} & 0.509 {\scriptsize[0.490,\,0.519]} \\
TIP-IM        & 0.496 {\scriptsize[0.496,\,0.501]} & 0.446 {\scriptsize[0.445,\,0.447]} & 0.944 {\scriptsize[0.943,\,0.945]} & 0.812 {\scriptsize[0.790,\,0.834]} & 0.878 {\scriptsize[0.868,\,0.889]} & 0.624 {\scriptsize[0.542,\,0.662]} & 0.607 {\scriptsize[0.593,\,0.618]} \\
Chameleon     & 0.239 {\scriptsize[0.235,\,0.246]} & 0.474 {\scriptsize[0.473,\,0.477]} & 0.965 {\scriptsize[0.964,\,0.966]} & 0.797 {\scriptsize[0.777,\,0.817]} & 0.881 {\scriptsize[0.870,\,0.892]} & 0.799 {\scriptsize[0.722,\,0.838]} & 0.508 {\scriptsize[0.503,\,0.516]} \\
\midrule
\multicolumn{8}{l}{\textit{(b) GAN-based generation}} \\
Adv-Makeup    & 0.012 {\scriptsize[0.012,\,0.012]} & 0.524 {\scriptsize[0.523,\,0.524]} & 0.967 {\scriptsize[0.967,\,0.968]} & 0.904 {\scriptsize[0.886,\,0.921]} & 0.936 {\scriptsize[0.927,\,0.944]} & 0.931 {\scriptsize[0.892,\,0.956]} & 0.056 {\scriptsize[0.055,\,0.057]} \\
AMT-GAN       & 0.156 {\scriptsize[0.156,\,0.162]} & 0.481 {\scriptsize[0.479,\,0.481]} & 0.938 {\scriptsize[0.937,\,0.939]} & 0.577 {\scriptsize[0.539,\,0.614]} & 0.758 {\scriptsize[0.738,\,0.778]} & 0.148 {\scriptsize[0.121,\,0.165]} & 0.282 {\scriptsize[0.261,\,0.296]} \\
CIAGAN        & 0.450 {\scriptsize[0.450,\,0.452]} & 0.276 {\scriptsize[0.275,\,0.276]} & 0.741 {\scriptsize[0.738,\,0.744]} & 0.575 {\scriptsize[0.546,\,0.602]} & 0.658 {\scriptsize[0.644,\,0.671]} & 0.247 {\scriptsize[0.193,\,0.270]} & 0.369 {\scriptsize[0.348,\,0.387]} \\
DeID-rPPG     & 0.013 {\scriptsize[0.013,\,0.013]} & 0.407 {\scriptsize[0.406,\,0.407]} & 0.942 {\scriptsize[0.940,\,0.943]} & 0.833 {\scriptsize[0.813,\,0.852]} & 0.887 {\scriptsize[0.877,\,0.897]} & 0.907 {\scriptsize[0.855,\,0.937]} & 0.061 {\scriptsize[0.059,\,0.061]} \\
G$^{2}$Face   & 0.210 {\scriptsize[0.209,\,0.213]} & 0.443 {\scriptsize[0.442,\,0.443]} & 0.903 {\scriptsize[0.902,\,0.905]} & 0.787 {\scriptsize[0.769,\,0.805]} & 0.845 {\scriptsize[0.836,\,0.854]} & 0.326 {\scriptsize[0.274,\,0.338]} & 0.400 {\scriptsize[0.384,\,0.406]} \\
\midrule
\multicolumn{8}{l}{\textit{(c) Diffusion-based generation}} \\
WeakenDiff    & 0.222 {\scriptsize[0.221,\,0.225]} & 0.371 {\scriptsize[0.370,\,0.371]} & 0.926 {\scriptsize[0.925,\,0.927]} & 0.643 {\scriptsize[0.610,\,0.674]} & 0.784 {\scriptsize[0.768,\,0.799]} & 0.185 {\scriptsize[0.151,\,0.200]} & 0.332 {\scriptsize[0.308,\,0.342]} \\
DiffAM        & 0.159 {\scriptsize[0.157,\,0.160]} & 0.471 {\scriptsize[0.471,\,0.472]} & 0.910 {\scriptsize[0.908,\,0.911]} & 0.700 {\scriptsize[0.674,\,0.725]} & 0.805 {\scriptsize[0.792,\,0.817]} & 0.557 {\scriptsize[0.455,\,0.611]} & 0.394 {\scriptsize[0.382,\,0.399]} \\
\bottomrule
\end{tabular}%
}
\end{table}

\paragraph{Findings.}
Tab.~\ref{tab:app_boot} lists the intervals. (i)~TIP-IM is top-1 in $100\%$ of the $2{,}000$ resamples; its margin over the runner-up is $0.097$ with 95\% CI $[0.086, 0.108]$. (ii)~The intervals of TI-DIM ($[0.490, 0.519]$) and Chameleon ($[0.503, 0.516]$) overlap and their order varies across resamples, so we report ranks 2 and 3 as a tie throughout the paper. (iii)~The image axes are estimated with half-widths below $0.006$, whereas the video axes are wider: the mean micro-expression interval width is $0.047$ against $0.0027$ for gaze (a ratio of $17.6$), and the PURE $U_3$ intervals are the widest in the table because ten subjects is few. This asymmetry is the reason L2 and L3 conclusions carry less evidential weight than L1. It does not put the L2 conclusions in doubt, since the intervals remain far apart for every comparison that matters (\eg, CIAGAN $U_2 = 0.658$ $[0.644, 0.671]$ against Adv-Makeup $0.936$ $[0.927, 0.944]$). (iv)~The small video corpora refine but do not drive the composite: scoring on the image axes alone ($\bar{P}$, $Q$, $U_1$, gaze, every small-$N$ video component removed) agrees with the full ranking at Spearman $0.888$ with the same top three, removing $Q$ leaves Spearman $0.979$, and the held-out MMPD replication (Appendix~\ref{sec:app_mmpd}) tightens the $U_3$ intervals by an order of magnitude while reproducing the PURE ordering. On the concern that the harmonic mean amplifies a noisy weak component: vetoing is intentional (a method that fails an axis should not rank highly), it acts on the \emph{mean} of each axis rather than on a single sample, and the intervals above make the residual uncertainty explicit.

\subsection{Estimator Substitution and Multi-Model Averaging}
\label{sec:app_estsub}
The consistency formulation transfers the annotation dependency to an estimator dependency. We therefore test, over every scored image in \uface, whether replacing an L1 estimator by an independently trained public checkpoint changes the ranking. The alternatives (listed with their architectures and training corpora in Tab.~\ref{tab:app_repro_weights}) are: for ethnicity, a ViT with its own \emph{five}-class label space, a stricter test than a same-taxonomy swap; for macro-expression, POSTER trained on AffectNet instead of RAF-DB, and an independently trained FER ViT; for age, a ViT age classifier instead of MiVOLO. We ran both direct substitution (one estimator swapped) and multi-model averaging. 

\begin{table}[h]
\centering
\caption{\textbf{Estimator substitution and averaging leave the ranking unchanged.} Sub-score correlation is the Spearman correlation between the per-method sub-score under the published estimator and under the alternative; \metric\ Spearman compares the resulting composite ranking with Tab.~\ref{tab:hifd_main}.}
\label{tab:app_estsub}
\resizebox{0.94\textwidth}{!}{%
\begin{tabular}{@{}llccc@{}}
\toprule
\textbf{Attribute} & \textbf{Substitution} & \textbf{Sub-score correlation} & \textbf{\metric\ Spearman} & \textbf{Rank changes} \\
\midrule
Ethnicity        & FairFace (7-class) $\to$ ViT ethnicity (5-class) & $0.937$ & $1.000$ & 0 \\
Macro-expression & POSTER/RAF-DB $\to$ POSTER/AffectNet           & $0.979$ & $1.000$ & 0 \\
Macro-expression & POSTER/RAF-DB $\to$ FER ViT                     & $0.979$ & $1.000$ & 0 \\
Age              & MiVOLO $\to$ ViT age                             & $0.944$ & $1.000$ & 0 \\
\midrule
All L1 sub-scores & averaged over 2--3 models each                  & --      & $1.000$ & 0 \\
\midrule
\multicolumn{5}{l}{\textit{Axes that already ensemble:} recognizers, pairwise Spearman $1.000$ / $0.965$ / $0.965$; rPPG estimators, $0.972$ / $1.000$ / $0.972$.} \\
\bottomrule
\end{tabular}%
}
\end{table}

Tab.~\ref{tab:app_estsub} summarizes: every substitution and the fully averaged composite give Spearman $1.000$ with the published ranking, zero rank changes, and the same top three. The alternatives also agree with the published estimators at the sub-score level ($0.937$ to $0.979$) despite different training corpora and label spaces, so the sub-scores themselves are not artefacts of one model. The averaged composite differs from Tab.~\ref{tab:hifd_main} by at most $0.002$ per method. Combined with the component-removal checks of Appendix~\ref{sec:app_uncertainty}, this bounds how much any single estimator can move the conclusions.

\subsection{Weight-Profile Sensitivity}
\label{sec:app_sweep}
\begin{figure}[h]
    \centering
    \includegraphics[width=0.95\linewidth]{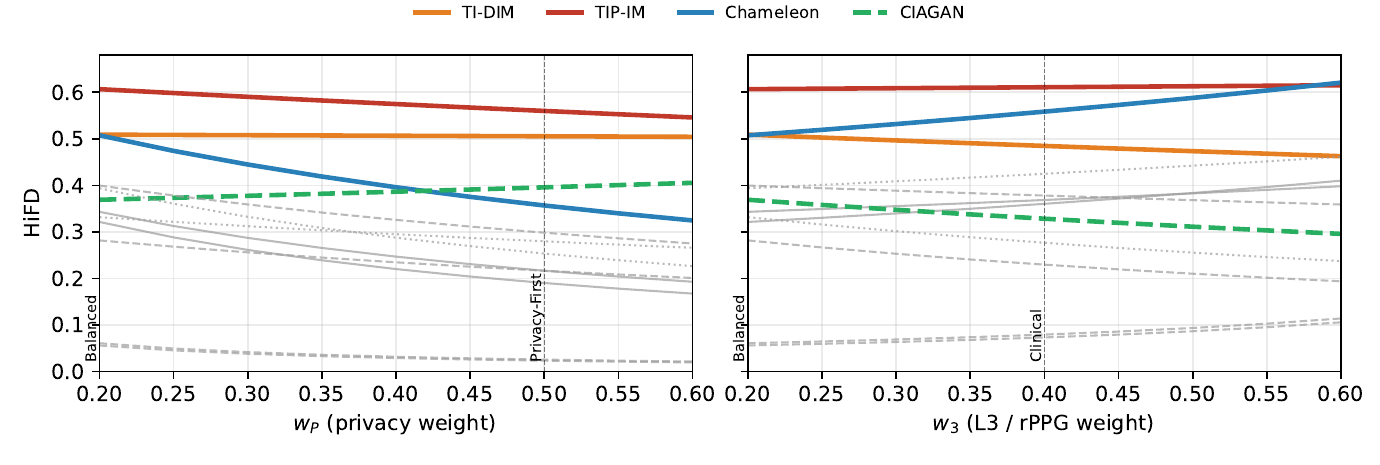}
    \vspace{-2mm}
    \caption{\textbf{Continuous weight sweeps.} \metric\ for all twelve methods as $w_P$ (left) or $w_3$ (right) varies from $0.2$ to $0.6$ with the remaining weight split equally; solid, dashed, and dotted lines are adversarial, GAN, and diffusion methods, and the four methods that ever enter the top three are highlighted.}
    \label{fig:weight_sweep}
\end{figure}

The three profiles of §\ref{sec:fdeidscore} are hand-specified. Fig.~\ref{fig:weight_sweep} re-aggregates the component scores of Tab.~\ref{tab:hifd_main} while one weight sweeps continuously from $0.2$ to $0.6$ and the remaining weight is split equally over the other four axes; Tab.~\ref{tab:app_sweep} tabulates the four methods that are ever in the top three. TIP-IM is rank~1 at every $w_P$; the only top-three change is CIAGAN, whose $\bar{P} = 0.45$ is the highest among GAN methods, rising from rank 6 to rank 3 as privacy is weighted more heavily. Under $w_3$, TIP-IM stays first until $w_3 = 0.6$, where Chameleon ($U_3 = 0.80$, the highest among methods with $\bar{P} > 0.2$) edges ahead by $0.006$. This $0.006$ margin is smaller than the half-width of either method's bootstrap interval at Balanced (Appendix~\ref{sec:app_uncertainty}), so the swap at the extreme of the sweep is a statistical tie rather than a reversal. The profiles are thus illustrative points on smooth curves, and the leader is not an artifact of any particular setting.

\begin{table}[h]
\centering
\caption{\textbf{Weight sweep, tabulated.} \metric\ of the four methods that enter the top three at any setting.}
\label{tab:app_sweep}
\resizebox{0.8\textwidth}{!}{%
\begin{tabular}{@{}l ccccc c ccccc@{}}
\toprule
& \multicolumn{5}{c}{$w_P$ (remainder equal)} & & \multicolumn{5}{c}{$w_3$ (remainder equal)} \\
\cmidrule(lr){2-6}\cmidrule(lr){8-12}
\textbf{Method} & 0.2 & 0.3 & 0.4 & 0.5 & 0.6 & & 0.2 & 0.3 & 0.4 & 0.5 & 0.6 \\
\midrule
TIP-IM    & \textbf{0.607} & \textbf{0.590} & \textbf{0.575} & \textbf{0.560} & \textbf{0.546} & & \textbf{0.607} & \textbf{0.609} & \textbf{0.611} & \textbf{0.613} & 0.615 \\
TI-DIM    & 0.509 & 0.508 & 0.507 & 0.506 & 0.504 & & 0.509 & 0.497 & 0.485 & 0.474 & 0.463 \\
Chameleon & 0.508 & 0.445 & 0.396 & 0.357 & 0.325 & & 0.508 & 0.532 & 0.558 & 0.588 & \textbf{0.621} \\
CIAGAN    & 0.369 & 0.378 & 0.386 & 0.396 & 0.406 & & 0.369 & 0.348 & 0.328 & 0.311 & 0.296 \\
\midrule
Leader    & \multicolumn{5}{c}{TIP-IM throughout} & & \multicolumn{4}{c}{TIP-IM} & Chameleon \\
\bottomrule
\end{tabular}%
}
\end{table}

\subsection{Quality Axis: Reference-Free Alternatives}
\label{sec:app_quality_ablation}
Half of $Q$ is LPIPS similarity to the original. Since successful identity replacement necessarily changes appearance, LPIPS could in principle conflate ``changed the face'' with ``degraded the face''. We tested this with two measures that never see the original, BRISQUE~\cite{mittal2011blind} and TOPIQ face-IQA~\cite{chen2024topiq}, on a paired $12{,}000$-image subset per method, and rebuilt a reference-free quality term $Q_{\text{nr}}$ as the mean of the two normalized scores. Tab.~\ref{tab:app_nrq} reports the results. Three observations. (i)~LPIPS is not the mechanism penalizing replacement methods: dropping it moves the replacement family \emph{down} on average (mean rank shift $-0.5$) and the perturbation family up ($+0.5$), and CIAGAN falls from rank $6$ to $9$ rather than rising, because its quality deficit is visible without any reference (BRISQUE $80.9$ and TOPIQ $0.250$ against $38.4$ and $0.752$ for the originals, and the highest face-detection failure rate of any method, $0.7\%$). Genuinely clean appearance-changing methods score near the originals (Adv-Makeup $39.0$, AMT-GAN $39.4$). (ii)~The published and reference-free $Q$ rank methods almost independently (Spearman $0.126$): the two capture different things, fidelity to the source versus freedom from artifacts, which is why we report the sub-scores separately. (iii)~The composite is insensitive to the choice: substituting $Q_{\text{nr}}$ leaves Spearman $0.951$ with the same top three, and removing $Q$ altogether leaves $0.979$. The human study (Appendix~\ref{sec:app_human}) adds that LPIPS has no detectable rank association with human naturalness, so $Q$ should be described as an artifact and fidelity term rather than a naturalness term.

\begin{table}[h]
\centering
\caption{\textbf{Reference-free quality and its effect on the composite.} BRISQUE (lower is better) and TOPIQ face-IQA (higher is better) on $12{,}000$ images per method (originals: $38.4$ / $0.752$). $Q_{\text{nr}}$ averages the normalized scores; ranks in parentheses are within the column. The last three columns are \metric\ with $Q_{\text{nr}}$ substituted, with $Q$ removed (four-axis harmonic mean), and as published.}
\label{tab:app_nrq}
\resizebox{0.9\textwidth}{!}{%
\begin{tabular}{@{}l cc cc ccc@{}}
\toprule
\textbf{Method} & \textbf{BRISQUE}$\downarrow$ & \textbf{TOPIQ}$\uparrow$ & $Q_{\text{nr}}$ (rank) & $Q$ (rank) & \metric[$Q_{\text{nr}}$] & \metric[no $Q$] & \metric \\
\midrule
\multicolumn{8}{l}{\textit{(a) Adversarial perturbation}} \\
MI-FGSM       & 15.7 & 0.633 & 0.738 (2) & 0.360 (10) & 0.380 (6) & 0.339 (7) & 0.343 (7) \\
PGD           & 14.9 & 0.648 & 0.749 (1) & 0.362 (9) & 0.354 (7) & 0.313 (9) & 0.321 (9) \\
TI-DIM        & 25.3 & 0.575 & 0.661 (4) & 0.342 (11) & 0.595 (2) & 0.580 (2) & 0.509 (2) \\
TIP-IM        & 40.5 & 0.670 & 0.632 (7) & 0.446 (5) & 0.660 (1) & 0.667 (1) & 0.607 (1) \\
Chameleon     & 40.5 & 0.701 & 0.648 (5) & 0.474 (3) & 0.538 (3) & 0.517 (3) & 0.508 (3) \\
\midrule
\multicolumn{8}{l}{\textit{(b) GAN-based generation}} \\
Adv-Makeup    & 39.0 & 0.740 & 0.675 (3) & 0.524 (1) & 0.056 (12) & 0.046 (12) & 0.056 (12) \\
AMT-GAN       & 39.4 & 0.666 & 0.636 (6) & 0.481 (2) & 0.290 (10) & 0.255 (10) & 0.282 (10) \\
CIAGAN        & 80.9 & 0.250 & 0.221 (12) & 0.276 (12) & 0.346 (9) & 0.403 (4) & 0.369 (6) \\
DeID-rPPG     & 69.7 & 0.497 & 0.400 (11) & 0.407 (7) & 0.061 (11) & 0.050 (11) & 0.061 (11) \\
G$^{2}$Face   & 48.6 & 0.632 & 0.573 (9) & 0.443 (6) & 0.417 (4) & 0.390 (5) & 0.400 (4) \\
\midrule
\multicolumn{8}{l}{\textit{(c) Diffusion-based generation}} \\
WeakenDiff    & 61.7 & 0.592 & 0.488 (10) & 0.371 (8) & 0.347 (8) & 0.323 (8) & 0.332 (8) \\
DiffAM        & 40.8 & 0.643 & 0.617 (8) & 0.471 (4) & 0.410 (5) & 0.378 (6) & 0.394 (5) \\
\midrule
\multicolumn{5}{l}{Spearman with published \metric} & 0.951 & 0.979 & -- \\
\bottomrule
\end{tabular}%
}
\end{table}

\section{Comparative Study: Extended Results}
\subsection{Qualitative Comparisons}

\begin{figure}[h]
    \centering
    \includegraphics[width=1.0\linewidth]{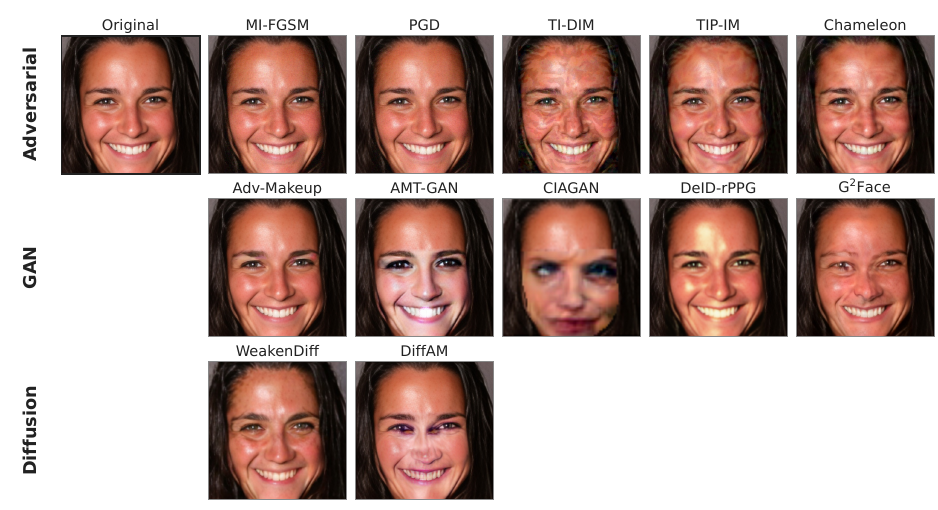}
    \vspace{-3mm}
    \caption{\textbf{Qualitative comparison on a representative \uface\ sample.} Original (top-left) and 12 de-identified outputs grouped by paradigm.}
    \label{fig:qualitative_grid0}
    \vspace{-3mm}
\end{figure}

\begin{figure}[h]
    \centering
    \includegraphics[width=1.0\linewidth]{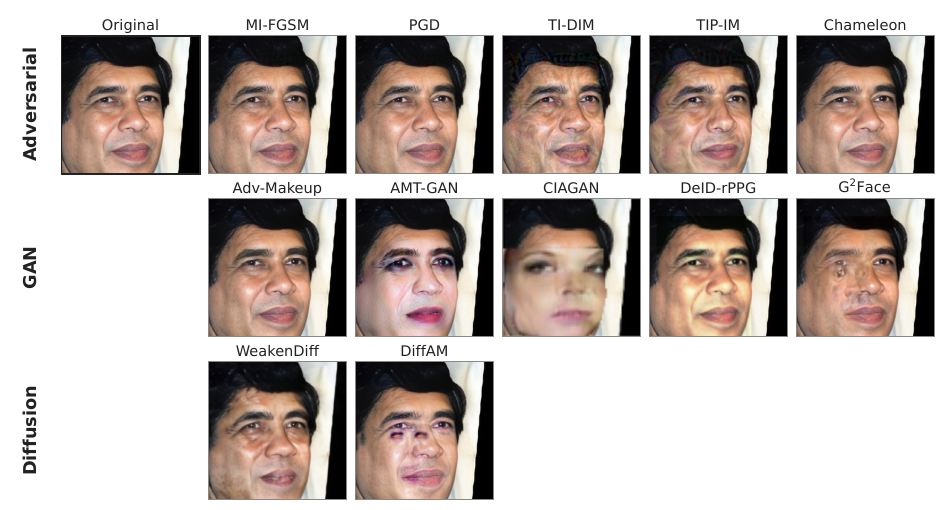}
    \vspace{-3mm}
    \caption{\textbf{Qualitative comparison on a representative \uface\ sample.} Same layout as Fig.~\ref{fig:qualitative_grid0}.}
    \label{fig:qualitative_grid1}
    \vspace{-3mm}
\end{figure}

We present qualitative comparisons on two representative \uface\ samples chosen to span pose and lighting variation: Fig.~\ref{fig:qualitative_grid0} shows a near-frontal female subject under uniform illumination, and Fig.~\ref{fig:qualitative_grid1} shows a three-quarter-view male subject under lateral lighting. The 12-method grids are organized identically (rows by paradigm, columns by method) to enable cross-method visual comparison.

\subsection{Demographic Fairness Under Profile Reweighting}
\label{sec:app_demographic_profiles}
\begin{figure}[h]
    \centering
    \includegraphics[width=1.0\linewidth]{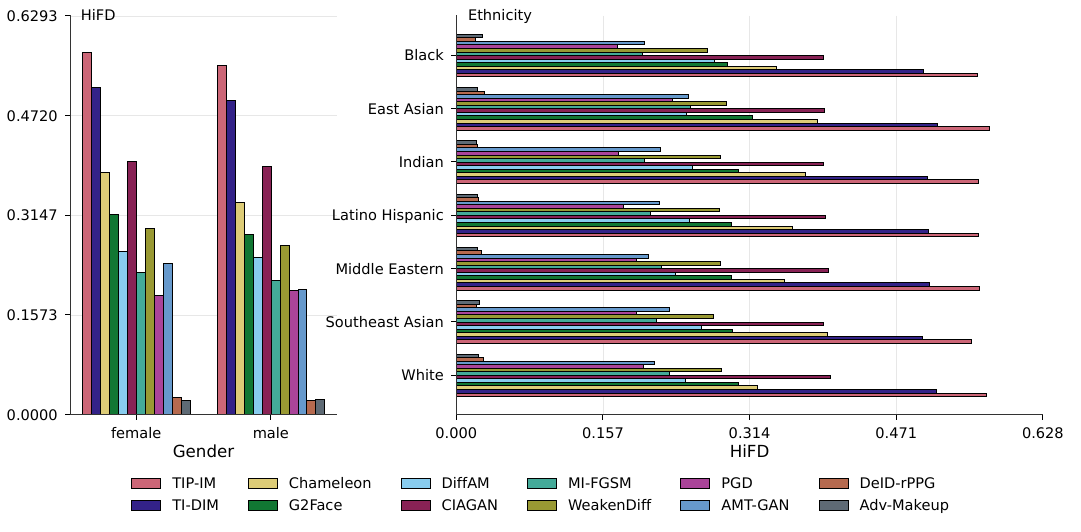}
    \vspace{-7mm}
    \caption{\textbf{Per-demographic-slice Privacy-First \metric\ on \uface.} Same layout as Fig.~\ref{fig:hifd_demographic}; methods are ordered by global Privacy-First \metric\ (descending). Left shows the female/male split and right shows the seven-way ethnicity split (alphabetical).}
    \label{fig:hifd_demographic_privacy_first}
    \vspace{-3mm}
\end{figure}

\begin{figure}[h]
    \centering
    \includegraphics[width=1.0\linewidth]{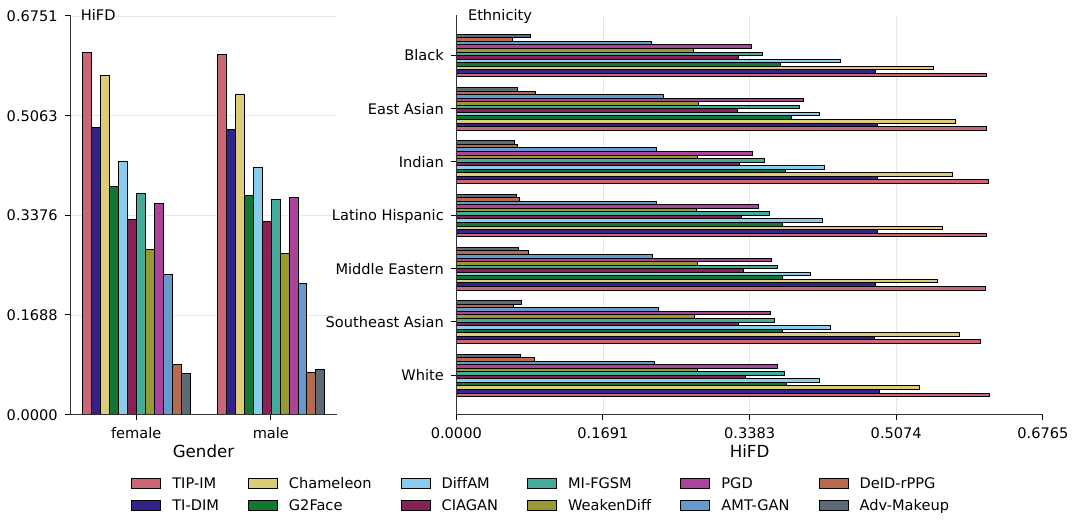}
    \vspace{-7mm}
    \caption{\textbf{Per-demographic-slice Clinical \metric\ on \uface.} Same layout as Fig.~\ref{fig:hifd_demographic}.}
    \label{fig:hifd_demographic_clinical}
    \vspace{-3mm}
\end{figure}

We extend the demographic decomposition of Fig.~\ref{fig:hifd_demographic} to the two non-balanced profiles defined in §\ref{sec:fdeidscore}: the \emph{Privacy-First} profile (Fig.~\ref{fig:hifd_demographic_privacy_first}), which doubles the privacy weight, and the \emph{Clinical} profile (Fig.~\ref{fig:hifd_demographic_clinical}), which emphasizes physiological L3 utility.

Under \emph{Privacy-First}, the aggregate male-female gap widens slightly to $0.018$ (female $0.287$, male $0.269$) and the ethnicity spread widens to $0.021$ (East Asian highest at $0.292$, Black lowest at $0.271$). Per-method, Chameleon shows the widest gender gap ($0.048$) and the widest ethnicity range ($0.075$, peaking on Southeast Asian), while CIAGAN, Adv-Makeup, and DeID-rPPG retain ranges below $0.010$.
Under \emph{Clinical}, the aggregate gaps shrink: $0.008$ for gender and $0.013$ for ethnicity, with PGD the widest by ethnicity ($0.059$, East Asian peak) and Chameleon the widest by gender ($0.032$).
Across both profiles, the qualitative structure observed on the Balanced profile is preserved: the East-Asian-peak / Black-trough ordering survives the reweighting on the aggregate. The per-method peak ethnicity is method-specific rather than profile-specific (PGD and MI-FGSM peak on East Asian, Chameleon on Southeast Asian, DiffAM on Black) and matches the Balanced ranking; and the methods with the smallest disparities under the Balanced \metric\ (TIP-IM, TI-DIM, CIAGAN, Adv-Makeup; ethnicity range $\leq 0.020$ under every profile) remain so.
The top-ranked method TIP-IM continues to lead under both profiles ($\metric\!=\!0.560$ Privacy-First, $0.611$ Clinical) with gender gap $\leq 0.021$ and ethnicity range $\leq 0.019$, so its leadership is not a profile-specific artifact.
Profile reweighting changes \emph{which} component is amplified, not \emph{which} subgroup is favored.

\subsection{L3 Replication on the MMPD Dataset}
\label{sec:app_mmpd}
PURE has ten subjects, so we replicate the L3 evaluation on a larger dataset. We de-identified the full MMPD dataset~\cite{tang2023mmpd} ($33$ subjects, $655$ trials, $392{,}832$ frames after intersecting the keys common to all methods; frame-identical across conditions) with all twelve methods and re-ran the three rPPG estimators with no retraining. We deliberately did not use UBFC-rPPG, on which PhysNet, EfficientPhys, and FactorizePhys are trained, to avoid a train/test leak. MMPD is stratified by four illumination conditions (LED-low, LED-high, incandescent, natural) and five motion conditions (stationary, stationary after exercise, rotation, talking, walking), which lets us check that the family-level separation is not an artifact of one recording condition.

\begin{table}[t]
\centering
\caption{\textbf{$U_3$ on MMPD versus PURE.} MMPD $U_3$; PURE $U_3$; the range of $U_3$ over the four illumination and five motion conditions of MMPD; and the contact-sensor validation of Appendix~\ref{sec:app_rppg_truth}: the increase in heart-rate MAE caused by de-identification ($\Delta$MAE, bpm) and the fraction of the original HR accuracy retained.}
\label{tab:app_mmpd}
\resizebox{\textwidth}{!}{%
\begin{tabular}{@{}l l c cc l c@{}}
\toprule
\textbf{Method} & $U_3$ \textbf{MMPD} {\scriptsize[95\% CI]} & $U_3$ \textbf{PURE} & \textbf{Illumination range} & \textbf{Motion range} & $\Delta$\textbf{MAE (bpm)} & \textbf{HR retention} \\
\midrule
\multicolumn{7}{l}{\textit{(a) Adversarial perturbation}} \\
MI-FGSM       & 0.503 {\scriptsize[0.491,\,0.514]} & 0.474 & [0.470, 0.520] & [0.400, 0.689] & 3.3 {\scriptsize[2.5,\,4.2]} & 0.834 \\
PGD           & 0.558 {\scriptsize[0.546,\,0.570]} & 0.566 & [0.523, 0.578] & [0.445, 0.755] & 2.2 {\scriptsize[1.5,\,3.0]} & 0.889 \\
TI-DIM        & 0.410 {\scriptsize[0.398,\,0.421]} & 0.424 & [0.379, 0.437] & [0.322, 0.585] & 4.1 {\scriptsize[3.2,\,5.0]} & 0.794 \\
TIP-IM        & 0.651 {\scriptsize[0.640,\,0.662]} & 0.624 & [0.614, 0.683] & [0.591, 0.767] & 1.7 {\scriptsize[1.0,\,2.4]} & 0.913 \\
Chameleon     & 0.709 {\scriptsize[0.700,\,0.718]} & 0.799 & [0.677, 0.723] & [0.677, 0.727] & 0.7 {\scriptsize[0.0,\,1.3]} & 0.966 \\
\midrule
\multicolumn{7}{l}{\textit{(b) GAN-based generation}} \\
Adv-Makeup    & 0.880 {\scriptsize[0.874,\,0.885]} & 0.931 & [0.870, 0.885] & [0.863, 0.920] & 0.0 {\scriptsize[$-$0.5,\,0.4]} & 1.001 \\
AMT-GAN       & 0.156 {\scriptsize[0.149,\,0.163]} & 0.148 & [0.149, 0.167] & [0.113, 0.232] & 7.3 {\scriptsize[6.3,\,8.3]} & 0.635 \\
CIAGAN        & 0.170 {\scriptsize[0.164,\,0.177]} & 0.247 & [0.169, 0.172] & [0.128, 0.224] & 7.4 {\scriptsize[6.5,\,8.4]} & 0.628 \\
DeID-rPPG     & 0.729 {\scriptsize[0.719,\,0.739]} & 0.907 & [0.684, 0.785] & [0.676, 0.816] & 0.6 {\scriptsize[$-$0.1,\,1.2]} & 0.971 \\
G$^{2}$Face   & 0.319 {\scriptsize[0.310,\,0.329]} & 0.326 & [0.293, 0.339] & [0.261, 0.434] & 3.6 {\scriptsize[2.7,\,4.5]} & 0.819 \\
\midrule
\multicolumn{7}{l}{\textit{(c) Diffusion-based generation}} \\
WeakenDiff    & 0.273 {\scriptsize[0.264,\,0.282]} & 0.185 & [0.256, 0.281] & [0.212, 0.429] & 4.5 {\scriptsize[3.6,\,5.4]} & 0.773 \\
DiffAM        & 0.441 {\scriptsize[0.432,\,0.449]} & 0.557 & [0.434, 0.446] & [0.378, 0.515] & 1.2 {\scriptsize[0.4,\,2.1]} & 0.939 \\
\midrule
\multicolumn{7}{l}{Spearman(MMPD, PURE) $= 0.986$ (Pearson $0.967$); same top three and bottom three. Spearman($U_3$, $-\Delta$MAE) $= 0.944$ on MMPD, $0.993$ on PURE.} \\
\bottomrule
\end{tabular}%
}
\end{table}

Tab.~\ref{tab:app_mmpd} reports the results. The PURE ordering replicates at Spearman $0.986$ with the same top three (Adv-Makeup, DeID-rPPG, Chameleon) and bottom three (AMT-GAN, CIAGAN, WeakenDiff), and the MMPD intervals (subject-level bootstrap over $33$ subjects) are five to six times tighter than the PURE intervals of Tab.~\ref{tab:app_boot}, so $U_3$ is no longer the small-sample component of the composite. Finding~2 replicates at the family level: the five adversarial methods average $U_3 = 0.566$ on MMPD ($0.577$ on PURE), the seven generative methods $0.424$ ($0.472$), and the five generative methods that resynthesize the face region (AMT-GAN, CIAGAN, G$^{2}$Face, WeakenDiff, DiffAM) only $0.272$ ($0.293$); the two GAN methods that leave the skin largely untouched, Adv-Makeup (eye-region makeup) and DeID-rPPG (trained to preserve the pulse), are the exceptions already noted. The separation is not an artifact of one condition: Adv-Makeup holds $U_3 \in [0.863, 0.920]$ across every illumination and motion condition and CIAGAN stays within $[0.128, 0.224]$ throughout. Two methods move noticeably between datasets (DeID-rPPG $0.907 \to 0.729$, Chameleon $0.799 \to 0.709$), consistent with MMPD's harder motion and lighting, without changing their ranks. The contact-sensor columns give the same ranking from ground truth rather than consistency.

\section{Human Perceptual Study}
\label{sec:app_human}
\metric\ is estimator-based. The study reported here asks whether its judgments \emph{agree} with human perception on the three axes where a mismatch would matter most: (H1)~does human naturalness track the reference-free quality terms rather than LPIPS; (H2)~does human-judged attribute preservation track $U_1$, with the concrete counter-hypothesis that CIAGAN preserves demographics to a human despite its low $U_1$; (H3)~does the privacy score track human identity matching. It is not a validation of \metric\ against human perception in general, and we report its limits with its results.

\subsection{Protocol}
\paragraph{Design.}
Within-subject, two parts, the unpaired part always first so that seeing an original cannot anchor the naturalness judgments. \emph{Part~1 (unpaired).} One face at a time, no original shown, condition not disclosed; two independent 5-point items: naturalness (``Could this be a real, unedited photograph?'', from \emph{Definitely edited} to \emph{Definitely a real photo}) and artifacts (``How visible are visual defects?'', from \emph{None visible} to \emph{Severe}). Unmodified originals are included as unlabeled stimuli. \emph{Part~2 (paired).} Original and de-identified image side by side with random left/right order; the rater ticks which of four attributes still match (age within about ten years, gender, ethnicity, facial expression) as independent checkboxes, then judges identity (same person / different people / unsure). Asking per attribute is deliberate: $U_1$ is an average of per-attribute consistency scores, so only a per-attribute human measurement is commensurable with it.

\paragraph{Stimuli.}
Thirty base identities drawn stratified across the $14$ gender-by-ethnicity strata of \uface, so the rated sample inherits the benchmark's balance; $13$ conditions ($12$ methods plus the unmodified original); $390$ unique images. Each rater sees a balanced block of $39$ single images ($3$ per condition), $24$ pairs ($2$ per method), and $4$ quality-control trials (two identical pairs, one deliberately corrupted image, and the originals themselves), about $15$ minutes in total.

\paragraph{Quality control, pre-specified.}
Three checks were fixed before collection: mean naturalness on originals at or above $3.0$; the corrupted image rated at most $2.0$; identical pairs judged ``same person''. A rater failing two or more checks is excluded; trials answered faster than $800$\,ms are dropped. $24$ raters were collected; the rule excluded $2$, leaving $22$ against a target of $30$. Check~1 turned out to be mis-specified against a restored-image benchmark (it fires for $10$ of $24$ raters because the originals themselves are rated only $3.32/5$; see below) and is retained only because it was pre-registered; every result below is stable under all four exclusion rules we tested, including one that keeps only $9$ raters.

\paragraph{Participants and consent.}
Participants were consenting adults ($\ge 18$) recruited as a convenience sample. Participation was voluntary.
No name, e-mail address, IP address, or other identifying information was recorded; responses stayed in the rater's browser until the rater chose to download and return a results file, with no server, cookies, or tracking, and withdrawal was possible at any point by closing the page. Raters were told that the faces are of real people from public research datasets and were asked not to save or share them. The consent screen and the two instruction screens are reproduced verbatim below.

\begin{quote}\small
\textbf{Face image perception study.} About 15 minutes. No personal data is collected.
\emph{What you will do.} You will look at photographs of faces and answer short questions about them. Some images are unmodified photographs; others have been altered by automatic privacy-protection software. Your job is to report what you see. There are no right or wrong answers on the rating questions.
\emph{About the images.} The faces are of real people, taken from public research datasets that are already used for scientific work under their stated licences. Please do not save, copy or share the images.
\emph{Your data.} No name, email, IP address or any identifying information is recorded. Responses stay in your browser until you download a results file at the end and return it to the researcher. No cookies and no tracking of any kind. You may stop at any time by closing this page; nothing is sent unless you choose to download and return the file. Responses are used only in aggregate, to evaluate an image-quality metric.
\emph{Before you start.} Use a laptop or desktop if you can, on a screen you can see clearly. Work somewhere without glare, at your normal viewing distance. If you use glasses or contacts for screen work, please wear them.
[\ \ ] I am at least 18 years old, I have read the above, and I agree to take part.

\textbf{Part 1 of 2: single images.} You will see one face at a time. For each one, answer two separate questions. \emph{Question 1. Is it a real photo?} Could this be an ordinary, unedited photograph of a person? Judge the image as a whole. Do not try to guess whether software was used, just report how real it looks to you. \emph{Question 2. Are there visible defects?} Blurring, smearing, warped features, visible seams or edges, odd colour patches, repeated texture. Report only what you can actually see. These two questions are independent. An image can look like a real photograph and still have a small defect somewhere, and a picture can be defect-free yet still look artificial. Please answer each on its own. You can click the options or press keys 1 to 5. Answer at a comfortable pace and trust your first impression; there is no time limit but you should not deliberate for long.

\textbf{Part 2 of 2: image pairs.} Now you will see two images side by side. One is an original photograph and the other has been altered by privacy software, but the order is random, so do not assume the original is on either side. \emph{Question 1. What still matches?} Tick each characteristic that looks the same in both images. Judge each one separately. It is perfectly normal for some to match and others not. \emph{Question 2. Same person?} Would you say these two images show the same individual? Choose Unsure if you genuinely cannot tell; that is a useful answer, not a failure. Judge only the characteristics asked about. The two images may differ in many other ways at the same time, and that is expected.
\end{quote}

\paragraph{Analysis, specified before collection.}
Per-condition means with cluster-bootstrap intervals that resample raters ($10{,}000$ resamples); method-level Spearman correlations of human naturalness against BRISQUE, TOPIQ, NIQE, LPIPS, and the published $Q$ (H1), of human attribute preservation against $U_1$ with per-method residuals (H2), and of human same-person rate against $\bar{P}$ and fixed-FAR success (H3), each with a permutation $p$-value over methods; Krippendorff's $\alpha$ for the ordinal items and Fleiss' $\kappa$ for the binary items. Reporting the residuals, not only the correlation, is what makes H2 a test rather than a demonstration.

\subsection{Results}

\begin{table}[h]
\centering
\caption{\textbf{Human ratings by condition ($22$ raters, $1{,}386$ trials).} Naturalness and artifacts are 5-point means on unpaired images (higher naturalness and lower artifacts are better); attributes kept is the fraction of the four attributes judged to match on pairs; same person is the fraction of pairs judged to show the same individual (lower is more private). Intervals are rater-level cluster bootstraps.}
\label{tab:app_human}
\resizebox{0.8\textwidth}{!}{%
\begin{tabular}{@{}l l c l l@{}}
\toprule
\textbf{Condition} & \textbf{Naturalness}$\uparrow$ & \textbf{Artifacts}$\downarrow$ & \textbf{Attributes kept}$\uparrow$ & \textbf{Same person}$\downarrow$ \\
\midrule
\textit{Unmodified original} & \textit{3.32} {\scriptsize[2.92,\,3.70]} & \textit{2.08} & -- & -- \\ \midrule
\multicolumn{5}{l}{\textit{(a) Adversarial perturbation}} \\
MI-FGSM       & 3.12 {\scriptsize[2.73,\,3.53]} & 2.08 & 0.966 {\scriptsize[0.898,\,1.000]} & 0.977 {\scriptsize[0.932,\,1.000]} \\
PGD           & 3.35 {\scriptsize[2.95,\,3.71]} & 2.02 & 0.932 {\scriptsize[0.847,\,0.994]} & 1.000 {\scriptsize[1.000,\,1.000]} \\
TI-DIM        & 1.32 {\scriptsize[1.14,\,1.52]} & 4.52 & 0.881 {\scriptsize[0.795,\,0.949]} & 0.886 {\scriptsize[0.773,\,0.977]} \\
TIP-IM        & 1.55 {\scriptsize[1.27,\,1.86]} & 4.29 & 0.858 {\scriptsize[0.761,\,0.938]} & 0.909 {\scriptsize[0.795,\,1.000]} \\
Chameleon     & 2.20 {\scriptsize[1.83,\,2.58]} & 3.32 & 0.938 {\scriptsize[0.869,\,0.989]} & 0.932 {\scriptsize[0.864,\,1.000]} \\
\midrule
\multicolumn{5}{l}{\textit{(b) GAN-based generation}} \\
Adv-Makeup    & 3.27 {\scriptsize[2.85,\,3.65]} & 2.11 & 0.972 {\scriptsize[0.920,\,1.000]} & 0.977 {\scriptsize[0.932,\,1.000]} \\
AMT-GAN       & 1.98 {\scriptsize[1.71,\,2.26]} & 3.58 & 0.830 {\scriptsize[0.739,\,0.909]} & 0.727 {\scriptsize[0.568,\,0.886]} \\
CIAGAN        & 1.20 {\scriptsize[1.02,\,1.42]} & 4.80 & \textbf{0.159} {\scriptsize[0.085,\,0.244]} & \textbf{0.114} {\scriptsize[0.000,\,0.250]} \\
DeID-rPPG     & 2.36 {\scriptsize[1.86,\,2.89]} & 3.24 & 0.903 {\scriptsize[0.818,\,0.966]} & 0.977 {\scriptsize[0.932,\,1.000]} \\
G$^{2}$Face   & 1.23 {\scriptsize[1.08,\,1.44]} & 4.56 & 0.722 {\scriptsize[0.614,\,0.824]} & 0.614 {\scriptsize[0.455,\,0.773]} \\
\midrule
\multicolumn{5}{l}{\textit{(c) Diffusion-based generation}} \\
WeakenDiff    & 2.26 {\scriptsize[1.91,\,2.64]} & 3.05 & 0.818 {\scriptsize[0.733,\,0.892]} & 0.795 {\scriptsize[0.682,\,0.909]} \\
DiffAM        & 1.15 {\scriptsize[1.03,\,1.32]} & 4.70 & 0.682 {\scriptsize[0.551,\,0.795]} & 0.636 {\scriptsize[0.455,\,0.795]} \\
\bottomrule
\end{tabular}%
}
\end{table}

\paragraph{Raters discriminated.}
Three methods (PGD, Adv-Makeup, MI-FGSM) are statistically indistinguishable from the unmodified original on naturalness (within-rater differences with intervals covering zero) and the other nine are all significantly below it (Tab.~\ref{tab:app_human}). Naturalness and artifacts correlate at $-0.85$ across trials, related but not redundant. The originals themselves score only $3.32/5$: \uface\ images pass through GFPGAN/CodeFormer restoration, so raters correctly decline to call them unedited photographs. This is a finding about the benchmark, not about the raters, and it means the naturalness ceiling of \uface\ is not $5$.

\paragraph{H1: what human naturalness tracks.}
At the method level, human naturalness correlates with BRISQUE at Spearman $0.517$ (permutation $p = 0.09$), with TOPIQ at $0.336$ ($p = 0.28$), with NIQE at $-0.133$, with the published $Q$ at $0.091$, and with LPIPS at exactly $0.000$ ($p = 1.00$). The defensible claim is the null: LPIPS similarity to the original has no detectable rank association with human naturalness. The reference-free alternatives do better (the paired gap BRISQUE minus LPIPS is $+0.49$ $[+0.32, +0.62]$ under the rater bootstrap), but with twelve methods even BRISQUE does not reach significance against method sampling, so we do not claim a validated naturalness proxy. \metric\ (Balanced) itself correlates at $-0.62$ with human naturalness; this is the intended trade-off rather than a disagreement, since naturalness runs at $-0.66$ against $\bar{P}$ and $+0.91$ against $U_1$, and within the human data the same-person rate correlates at $+0.89$ with naturalness: the methods humans cannot see through are the ones they rate unnatural. \metric\ does not, and should not, track naturalness.

\paragraph{H2: attribute preservation and the CIAGAN question.}
Human attribute preservation tracks $U_1$ at Spearman $0.762$ $[0.567, 0.848]$, permutation $p = 0.007$. The counter-hypothesis that CIAGAN may preserve recognizable demographic attributes to a human despite its low $U_1$ is refuted in the opposite direction: humans put CIAGAN's attribute preservation at $0.159$ $[0.085, 0.244]$, far \emph{below} its $U_1$ of $0.468$, with $\Pr(\text{human} \ge U_1) = 0.000$ under the rater bootstrap. If $U_1$ errs on CIAGAN, it errs by being too generous. The largest positive residual is TI-DIM, where humans credit more preservation ($0.881$) than $U_1$ does ($0.757$).

\paragraph{H3: privacy, and why pooling is the wrong test.}
Pooled over all twelve methods, the human same-person rate correlates with $\bar{P}$ at $-0.549$ ($p = 0.07$) and with fixed-FAR success at only $-0.247$. This pools two threat models. Within the five adversarial methods the Spearman correlation with fixed-FAR success is $-1.000$, and within the seven generative methods (GAN-based and diffusion-based) it is $-1.000$: humans reproduce the operational de-identification ranking exactly inside each paradigm. The pooled figure is low because adversarial methods are engineered to defeat a recognizer while remaining near-invisible to a person (PGD same-person rate $1.000$ at $\bar{P} = 0.113$), whereas replacement methods change the face outright (CIAGAN $0.114$). The correct statement is that $\bar{P}$ is a machine-recognizer privacy score, it ranks operational risk correctly within a paradigm, and humans are not the adversary it models.

\paragraph{The frontier, measured entirely by humans.}
No method hides identity from a human (same-person rate $< 0.5$) while preserving attributes to a human ($> 0.7$). Only CIAGAN clears the identity bar, and it destroys the attributes doing so; every method that keeps attributes above $0.7$ is still recognized as the same person by at least $61\%$ of raters. Sweeping the identity threshold over $\{0.3, 0.4, 0.5, 0.6\}$ and the attribute threshold over $\{0.5, 0.6, 0.7, 0.8\}$ leaves the set empty in all sixteen combinations; only at an identity threshold of $0.7$, meaning $70\%$ of raters still say ``same person'', does any method qualify, which is not de-identification. 

\paragraph{Reliability and limitations.}
Krippendorff's $\alpha$ is $0.42$ on naturalness and $0.64$ on artifacts; Fleiss' $\kappa$ on the attribute majority is $0.45$. The naturalness value is below the conventional $0.667$ and must be read as such; two contributors are uneven block assignment (raters were not assigned round-robin, so $78$ of $312$ (identity, condition) cells are singly rated and one block drew no rater) and the compressed effective range (originals at $3.32$ rather than at ceiling), which deflates $\alpha$ mechanically. Method-level correlations rest on twelve points, and the permutation $p$-values are the conservative reading. Raters are a convenience sample rather than a demographically balanced panel. Every conclusion above survives all four exclusion rules (pre-specified, $n = 22$; none, $n = 24$; dropping the mis-specified check, $n = 18$; any single failure excludes, $n = 9$). For these reasons the study supports three specific claims (LPIPS is not what humans respond to; $U_1$ is not too generous to attribute-destroying methods; $\bar{P}$ is a recognizer-side score) and one replication of the central finding, not a general validation of \metric\ against human perception.

\section{Reproducibility Details}
\label{sec:app_repro}

\subsection{Hardware and Software Environment}

All inference and scoring jobs were executed on a supercomputer equipped with AMD MI250X GPUs. We use a Conda environment built on \texttt{rocm/6.0.3} with PyTorch~$2.3.1$. The single global random seed is $42$, applied to NumPy, PyTorch (CUDA/HIP RNGs), and Python's \texttt{random} module at the top of every extraction script. All estimators are run in inference mode with a fixed batch order.

\subsection{Pretrained Estimator Checkpoints}
\label{sec:app_repro_weights}

Tab.~\ref{tab:app_repro_weights} lists every pretrained checkpoint used in \metric.
All checkpoints are released by their original authors under research-only licenses.

\begin{table}[t]
\centering
\caption{\textbf{Pretrained checkpoints used in \metric.} URLs as released by the original authors.
\dag~Architecture is the inference backbone. ArcFace, CosFace, and AdaFace are insightface-style IR networks trained with different
losses on different data.}
\label{tab:app_repro_weights}
\resizebox{\textwidth}{!}{%
\begin{tabular}{@{}llllr@{}}
\toprule
\textbf{Component} & \textbf{Model} & \textbf{Arch.\dag} & \textbf{Train data} & \textbf{Source} \\
\midrule
\multicolumn{5}{l}{\textit{Privacy ensemble}} \\
$\bar{P}$ (1/3) & ArcFace~\cite{deng2019arcface}        & IR-R100  & MS1MV3       & insightface model zoo \\
$\bar{P}$ (2/3) & CosFace~\cite{wang2018cosface}        & IR-R50   & Glint360K    & insightface model zoo \\
$\bar{P}$ (3/3) & AdaFace~\cite{kim2022adaface}         & IR-50    & MS1MV2       & official release \\
\midrule
\multicolumn{5}{l}{\textit{L1 macro-cues}} \\
Age, Gender     & MiVOLO~\cite{kuprashevich2023mivolo}   & ViT-MiVOLO & UTKFace      & official release \\
Ethnicity       & FairFace~\cite{karkkainen2021fairface} & ResNet-34  & FairFace 7-race & official release \\
Macro-expression& POSTER~\cite{zheng2023poster}          & POSTER     & RAF-DB ($K{=}7$) & official release \\
Landmarks (68)  & MediaPipe~\cite{lugaresi2019mediapipe} & FaceLandmarker & in-the-wild faces & Google MP tasks \\
\midrule
\multicolumn{5}{l}{\textit{L1 alternatives used only in the substitution study (Appendix~\ref{sec:app_estsub})}} \\
Age (alt.)      & ViT age classifier                     & ViT-B/16   & FairFace age bins & HF \texttt{nateraw/vit-age-classifier} \\
Ethnicity (alt.)& ViT ethnicity (5-class)                & ViT-B/16   & independent 5-class corpus & HF \texttt{cledoux42/Ethnicity\_Test\_v003} \\
Macro-exp.\ (alt.)& POSTER~\cite{zheng2023poster}        & POSTER     & AffectNet ($K{=}7$) & official release \\
Macro-exp.\ (alt.)& FER ViT                              & ViT-B/16   & FER-2013 ($K{=}7$) & HF \texttt{trpakov/vit-face-expression} \\
\midrule
\multicolumn{5}{l}{\textit{L2 micro-cues}} \\
Gaze            & L2CS-Net~\cite{abdelrahman2023l2cs}    & ResNet-50  & Gaze360~\cite{kellnhofer2019gaze360} & official release \\
Micro-expression& SAMER~\cite{ruan2022mimicking}         & ResNet-18 ($\times 2$) & CASME~II ($K{=}5$, 26 LOSO folds) & official release \\
\midrule
\multicolumn{5}{l}{\textit{L3 imperceptible cues}} \\
rPPG (1/3)      & PhysNet~\cite{yu2019remote}            & 3D-CNN     & UBFC-rPPG    & rPPG-Toolbox~release \\
rPPG (2/3)      & EfficientPhys~\cite{liu2023efficientphys} & TS-CNN     & UBFC-rPPG    & rPPG-Toolbox~release \\
rPPG (3/3)      & FactorizePhys~\cite{joshi2024factorizephys} & FSAM     & UBFC-rPPG    & rPPG-Toolbox~release \\
\midrule
\multicolumn{5}{l}{\textit{Quality}} \\
LPIPS-VGG       & LPIPS~\cite{zhang2018unreasonable}     & VGG-16     & --        & \texttt{pyiqa} \\
NIQE            & NIQE~\cite{mittal2012making}           & hand-crafted & --         & \texttt{pyiqa} \\
\midrule
\multicolumn{5}{l}{\textit{Detection / alignment (cached once on the original)}} \\
Face detection  & RetinaFace~\cite{deng2020retinaface}   & ResNet-50  & WIDER FACE   & official release \\
\bottomrule
\end{tabular}%
}
\end{table}


\end{document}

%% file: style.tex
\usepackage[T1]{fontenc}
\usepackage{graphicx}
\usepackage{comment}
\usepackage{amsmath,amssymb} %
\usepackage{color}
\usepackage{xcolor}
\usepackage{booktabs}
\usepackage{subcaption}
\usepackage[colorlinks=true]{hyperref}
\usepackage{bm}
\usepackage{xspace}
\usepackage{dsfont}
\usepackage{graphics}

\usepackage{textcomp}
\usepackage{microtype}

\usepackage[all]{nowidow}

\usepackage{algorithm,algorithmic}
\usepackage{wrapfig,epsfig}
\usepackage{mathtools}
\usepackage{dsfont}
\usepackage{graphicx}
\usepackage{nomencl}
\usepackage{multirow}
\usepackage{makecell}
\usepackage{adjustbox}
\usepackage{subcaption}
\usepackage{enumerate}
\usepackage{enumitem}
\usepackage{multirow}
\usepackage{amssymb}
\usepackage[symbol]{footmisc}
\usepackage{colortbl}
\usepackage{diagbox}
\usepackage{fontawesome}

\usepackage{booktabs}
\usepackage{bbding}
\usepackage{pifont}
\usepackage{wasysym}
\usepackage{utfsym}

\usepackage{xcolor}

\usepackage{hyperref}
\definecolor{modernCite}{RGB}{0, 160, 140}
\definecolor{modernLink}{RGB}{140, 0, 210}  
\definecolor{modernURL}{RGB}{220, 100, 0}   

\hypersetup{
    colorlinks=true,         
    citecolor=modernCite,   
    linkcolor=modernLink,   
    urlcolor=modernURL,   
    pdftitle={My Document},  
}

\usepackage{tcolorbox}
\tcbuselibrary{minted,skins}
\usepackage{float}

\definecolor{codegreen}{rgb}{0.0, 0.5, 0.4}
\definecolor{codegray}{rgb}{0.5, 0.5, 0.5}
\definecolor{codepurple}{rgb}{0.58, 0, 0.82}
\definecolor{codeblue}{rgb}{0.0, 0.4, 0.7}
\definecolor{codered}{rgb}{0.7, 0.1, 0.1}
\definecolor{backcolour}{rgb}{1.0, 1.0, 1.0}

\usepackage{listings}
\lstdefinestyle{pythonstyle}{
    backgroundcolor=\color{backcolour},
    commentstyle=\color{codegreen},
    keywordstyle=\color{codeblue}\bfseries,
    stringstyle=\color{codered},
    basicstyle=\ttfamily\small,
    breakatwhitespace=false,
    breaklines=true,
    captionpos=t,
    keepspaces=true,
    showspaces=false,
    showstringspaces=false,
    showtabs=false,
    tabsize=4,
    frame=none,
    xleftmargin=2mm,
    xrightmargin=2mm,
    aboveskip=0pt,
    belowskip=0pt,
    emph={self, True, False, None},
    emphstyle=\color{codeblue},
    morekeywords={def, class, return, raise, for, in, if, else, import, from, as, with},
}
\newcommand{\metric}{\textbf{HiFD}}
\newcommand{\uface}{\textsc{UtilFace}}
\newcommand{\cmark}{\ding{51}}
\newcommand{\xmark}{\ding{55}}

\newcommand{\eg}{\textit{e.g.}}